\PassOptionsToPackage{shortlabels}{enumitem}
\documentclass[11pt, a4paper]{include/gdm_format}
\usepackage[authoryear, sort&compress, round]{natbib}

\usepackage{xcolor}
\usepackage{subcaption}
\usepackage{mwe}
\usepackage{graphicx}
\usepackage{xr}
\usepackage{marvosym}
\usepackage{makecell}
\usepackage{fvextra}

\definecolor{bgcolor}{rgb}{0.95,0.95,0.95}

\usepackage{tablefootnote}
\usepackage{pdfpages}
\usepackage{etoc}

\usepackage{hyperref}
\usepackage{url}
\usepackage{xurl}
\usepackage[parfill]{parskip}
\usepackage{multirow}
\usepackage{multicol}
\usepackage{adjustbox}
\usepackage{colortbl}
\usepackage[table]{xcolor}
\usepackage{svg}

\usepackage{microtype}
\usepackage{booktabs} % for professional tables

\usepackage{hyperref}

\usepackage{amsmath}
\usepackage{amssymb}
\usepackage{mathtools}
\usepackage{amsthm}

\usepackage[capitalize,noabbrev]{cleveref}

\usepackage{array}
\usepackage{ragged2e}
\usepackage{enumitem}
\usepackage[dvipsnames]{xcolor}

\usepackage{tablefootnote}
\usepackage{pdfpages}
\usepackage{etoc}

\usepackage{sidecap}
\usepackage{soul}
\sidecaptionvpos{figure}{t}
\usepackage{placeins}

\usepackage[linesnumbered,ruled,vlined]{algorithm2e}

\newcommand{\mb}[1]{\mathbf{#1}}

\newcommand{\implname}{FwPKM}

\definecolor{slowBlue}{HTML}{D0DAEE}
\definecolor{fastGreen}{HTML}{D5E8D4}
\definecolor{fastRed}{HTML}{F8CECC}

\definecolor{borderGreen}{HTML}{82B366} % Darker Green for Addressing border
\definecolor{borderRed}{HTML}{B85450}   % Darker Red for Memorization border

\newcommand{\cbox}[2]{\setlength{\fboxsep}{1.5pt}\colorbox{#1}{#2}}

\newcommand{\lossbox}[3]{%
  \setlength{\fboxsep}{1.5pt}% Padding inside box
  \setlength{\fboxrule}{0.8pt}% Thickness of the border
  \fcolorbox{#1}{#2}{#3}%
}

\SetAlFnt{\small} % Sets the algorithm font size to 'small' (try \footnotesize for smaller)

\SetCommentSty{mycommfont} % Apply style to comments

\SetKwSty{mykwfont} % Apply style to keywords (Input, Output, For, etc.)
\definecolor{commentgreen}{RGB}{34, 139, 34}
\definecolor{keywordblue}{RGB}{0, 51, 153}

\title{Fast-weight Product Key Memory}

\correspondingauthor{Tianyu Zhao (tianyu@sakana.ai)}

\author[1]{Tianyu Zhao}
\author[1]{Llion Jones}

\affil[1]{Sakana AI}

\begin{document}

\begin{abstract}
Sequence modeling layers in modern language models typically face a trade-off between storage capacity and computational efficiency. While softmax attention offers unbounded storage at prohibitive quadratic cost, linear variants are more efficient but suffer from limited, fixed-size storage. We introduce Fast-weight Product Key Memory (\implname), a sparse fast-weight memory layer that resolves this tension. \implname~updates sparsely activated parameters at both training and inference time using chunk-level gradient descent on a local memory-rewrite objective. This performs Test-Time Training (TTT)-style gradient updates on activated slots in a sparse memory, enabling rapid memorization and retrieval of many new key–value associations while keeping per-token compute low and fixed. Experiments show that \implname~functions as an effective episodic memory that complements the semantic memory of standard modules, yielding significant perplexity reductions on long-context datasets. Notably, in Needle-in-a-Haystack evaluations, \implname~generalizes to 128K-token contexts despite being trained on only 4K-token sequences.
\end{abstract}

\maketitle

\vspace{0.5em}
\begin{center}
    \begin{tabular}{rcl}
        \raisebox{-1.5pt}{\includegraphics[height=1.05em]{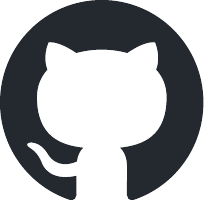}} & \textbf{Code} & \href{https://github.com/SakanaAI/fast-weight-product-key-memory}{\path{https://github.com/SakanaAI/fast-weight-product-key-memory}} \\
    \end{tabular}
\end{center}
\vspace{0.5em}

\setcounter{tocdepth}{1}
\etocdepthtag.toc{mtchapter}
\etocsettagdepth{mtchapter}{subsection}
\etocsettagdepth{mtappendix}{none}
\tableofcontents
\clearpage

\section{Introduction}  \label{sec:intro}

Long-context language modeling often requires remembering specific, local facts (\textit{e.g.} a name, a variable binding, a constraint) introduced tens of thousands of tokens earlier, and retrieving them when needed. The core sequence modeling layers, or token mixers, can be viewed as \textit{associative memory} systems~\citep{Transformer,ParallelDeltaNet,GDN,RWKV7,Mamba,Mamba2,TTT,Titans}. In this framework, information from past tokens is encoded into storage (\textit{i.e.} the process of \textit{memorization}) and later integrated into a prediction step by querying that storage (\textit{i.e.} the process of \textit{retrieval}).

The current landscape is defined by a trade-off between memory capacity and computational efficiency. Softmax self-attention~\citep{Transformer,TransformerAssocMem} \textit{memorizes} by explicitly storing a key-value pair for every past token and \textit{retrieves} by comparing the current query to all stored keys, producing a weighted sum of values. This yields high-fidelity retrieval but at quadratic cost with sequence length. In contrast, modern Recurrent Neural Networks (RNNs) such as linear attention~\citep{LinearAttention}, Mamba2~\citep{Mamba2}, and DeltaNet~\citep{FWLinearAttention,ParallelDeltaNet} compress history into a fixed-size state using learned update rules (\textit{e.g.}, Hebbian-style accumulation or Delta-rule error correction) and retrieve via a state readout in the form of vector–matrix multiplication. While this enables efficient, sub-quadratic retrieval, the fixed state capacity inherently limits the depth of information the model can retain compared to softmax attention.

To address the capacity limitations of these modern RNNs, recent work on Test-Time Training (TTT,~\citealt{TTT,TTTPlus}) and Titans~\citep{Titans} replaces the fixed state matrix with a ``fast-weight''~\citep{FW-Hinton,FW-Schmidhuber} neural network (typically an MLP). Similar to a state matrix, this MLP adapts to input sequences on-the-fly by minimizing a reconstruction loss. However, these approaches face a scaling bottleneck: dense network structure. An MLP’s computation scales linearly with its parameter count. To store the massive amount of information required for long-context tasks, the fast-weight MLP must be large; yet, updating and querying a large, dense MLP frequently is computationally infeasible.

In this paper, we show that sparse associative memory offers a way around this constraint. Sparse memory architectures such as Product Key Memory (PKM,~\citealt{PKM}) can maintain a very large number of memory parameters while activating only a small subset per token. We introduce Fast-weight Product Key Memory (\implname), which transforms PKM from a static, slow-weight module into a dynamic fast-weight episodic memory. By redesigning the sparse memory to update its key/value matrices via gradient descent at inference time, we enable \implname~to memorize vast amounts of context with fixed per-token compute. This distinction allows us to conceptually separate memory roles: standard slow weights serve as \textit{Semantic Memory} (storing dataset-wide facts), while \implname~serves as \textit{Episodic Memory} (storing context-specific bindings).

Our main contributions are threefold. 
\begin{itemize}[nosep]
    \item \textbf{Architecture.} We introduce Fast-weight Product Key Memory (\implname), a new sequence modeling layer that enables ``Sparse Test-Time Training'' -- \textit{TTT}-style gradient updates over a \textit{sparse} memory. We detail the architectural modifications and optimization objectives required to make sparse fast weights stable and effective (Section~\ref{sec:fwpkm}).
    \item \textbf{Performance.} We demonstrate that \implname~acts as a robust episodic memory when plugged into hybrid backbones. In Needle-in-a-Haystack evaluations, \implname~generalizes to 128K-token contexts despite being trained on only 4K sequences, with iterative reading boosting retrieval accuracy from $<10\%$ to $>70\%$ (Section~\ref{ssec:niah_eval}). \implname~also consistently reduces perplexity on long-context datasets like LC64 and Longbench (Section~\ref{ssec:ppl_eval},~\ref{ssec:longbench_eval}). Furthermore, we analyze continual learning capabilities, finding that while the model adapts to new domains online, it faces challenges with retention, motivating future work in memory consolidation (Section~\ref{ssec:pile_domain_eval}).
    \item \textbf{Analysis.} We provide interpretability analyses showing that \implname~explicitly captures informative entities and novel patterns (Section~\ref{sec:interpret}), alongside a cost analysis highlighting where sparse memory helps and where faster kernels are needed for broader scaling (Section~\ref{sec:cost}).
\end{itemize}

% \begin{figure}[t]
%     \begin{center}
%         \includesvg[width=1.0\textwidth]{figs/FwPKM_flow.svg}
%         \caption{An \implname~layer maintains key and value matrices as fast weights and update them after a chunk of tokens (512 tokens in this work). For a 4096-token input sequence, the fast weights are updated 8 times. \label{fig:fwpkm_flow}}
%     \end{center}
% \end{figure}

\section{Preliminary} \label{sec:prelim}

We briefly review sparse key-value memories and Product Key Memory (PKM,~\citealt{PKM}), which we later extend into an online-updated fast-weight module.

\subsection{Sparse Top-$k$ key-value memory}  \label{ssec:sparse_mem}
A standard key-value memory consists of a key matrix $K \in \mathbb{R}^{N \times d_k}$ and a value matrix $V \in \mathbb{R}^{N \times d_v}$, where $N$ is the number of memory slots and $d_{\{k,v\}}$ are the hidden dimensions.
Given a query vector $\mb{q} \in \mathbb{R}^{d_k}$, the model scores each slot (\textit{e.g.}, via dot product), 
\begin{align}
s_i &= \mb{q}^\top K_i,
\end{align}
selects the indices of the Top-$k$ scores $\mathcal{I}$,
\begin{align}
\mathcal{I} &= \texttt{Top-}k(\{s_i\}_{i=1}^N).
\end{align}
Then it computes normalized scores $\{s^\prime_i\}$ via softmax,
\begin{align}
\{s^\prime_i\} &= \mathtt{softmax}(\{s_j\}_{j \in \mathcal{I}}),
\end{align}
and retrieves a weighted sum of values,
\begin{align}
\hat{\mb{v}} &= \textstyle \sum_{i \in \mathcal{I}} s^\prime_i V_i.
\end{align}
Top-$k$ sparsity reduces the number of \textit{accessed} value rows, but still requires \textit{scoring} all $N$ keys to identify the top candidates. This linear $O(N)$ complexity prohibits scaling to massive memory sizes (\textit{e.g.} $N \approx 10^6$).

\subsection{Product Key Memory (PKM)}  \label{ssec:pkm}
Product Key Memory~\citep{PKM} reduces the \textit{scoring} cost by factorizing the keys into two smaller codebooks. PKM splits the query into two sub-queries $\mb{q} = [\mb{q}^{(1)}; \mb{q}^{(2)}]$, where $\mb{q}^{(1)}, \mb{q}^{(2)} \in \mathbb{R}^{d_k/2}$, and maintains two sub-key matrices $K^{(1)}, K^{(2)} \in \mathbb{R}^{\sqrt{N} \times d_k/2}$. It scores each codebook independently,
\begin{align}
s^{(1)}_i &= \mb{q}^{(1)\top} K^{(1)}_i,\quad
s^{(2)}_j = \mb{q}^{(2)\top} K^{(2)}_j,  \label{equ:pkm_score}
\end{align}
selects Top-$k$ indices for each codebook,
\begin{align}
\mathcal{I}^{(1)} = \texttt{Top-}k(\{s^{(1)}_i\}_{i=1}^{\sqrt{N}}),
\quad 
\mathcal{I}^{(2)} = \texttt{Top-}k(\{s^{(2)}_i\}_{i=1}^{\sqrt{N}}),
\end{align}
and then searches over the restricted Cartesian product $\mathcal{I}^{(1)} \times \mathcal{I}^{(2)}$ using additive pairwise scores
\begin{align}
s_{i,j} &= s^{(1)}_i + s^{(2)}_j,\quad (i,j)\in \mathcal{I}^{(1)} \times \mathcal{I}^{(2)}.
\end{align}
Finally, PKM selects the Top-$k$ pairs from this restricted set, applies softmax over the selected pair scores, and retrieves from the corresponding rows of the value table $V \in \mathbb{R}^{N \times d_v}$, where 1-indexed pair $(i,j)$ maps to row $((i-1)\times\sqrt{N} + j)$.

This decomposition of key space yields an efficient Top-$k$ retrieval that avoids scoring all $N$ slots. It allows PKM to index millions of slots (\textit{e.g.}, $10^6$) using only thousands of scores ($2\times 10^3$), making it an ideal choice for high-capacity associative memory.

% \begin{table*}[h]
% \centering
% \caption{Comparison of PKM and \implname}
% \label{tab:comp_pkm_\implname}
% % \renewcommand{\arraystretch}{1.6} % Adds breathing room between rows
% % \setlength{\tabcolsep}{12pt}      % Adds space between columns

% \begin{tabular}{@{} >{\RaggedRight\bfseries}p{0.2\textwidth} 
%                     >{\RaggedRight}p{0.35\textwidth} 
%                     >{\RaggedRight}p{0.35\textwidth} @{}}
% \toprule

% & \textbf{PKM} & \textbf{\implname} \\
% \midrule

% \textbf{Weight Type} & Slow Weights & Fast Weights \\ 
% \midrule

% Similar Modules & 
% FFN, MoE-FFN, PKM & 
% Softmax Attn., Linear Attn., DeltaNet \\ 
% \midrule

% Role & 
% \textbf{Channel Mixer}. Mixes features within a single token representation. & 
% \textbf{Token Mixer}. Mixes information across time steps (sequence positions). \\ 
% \midrule

% Parameter Update & 
% Updated at training; Frozen at inference. & 
% Updated at both training and inference. \\ 
% \midrule

% Memory Horizon & 
% \textbf{Long-term (Semantic)}. Stores dataset-wide facts and general rules (e.g., world knowledge). & 
% \textbf{Short-term (Episodic)}. Stores context-specific variable bindings (the Context Window). \\ 
% \midrule

% Learning Objective & 
% \textbf{Global Objective} of next token prediction & 
% \textbf{Local Objective} of memorization \\ 

% \bottomrule
% \end{tabular}
% \end{table*}
\section{Fast-weight Product Key Memory} \label{sec:fwpkm}

\begin{figure}[t!]
    \begin{center}
        \includesvg[width=0.75\textwidth]{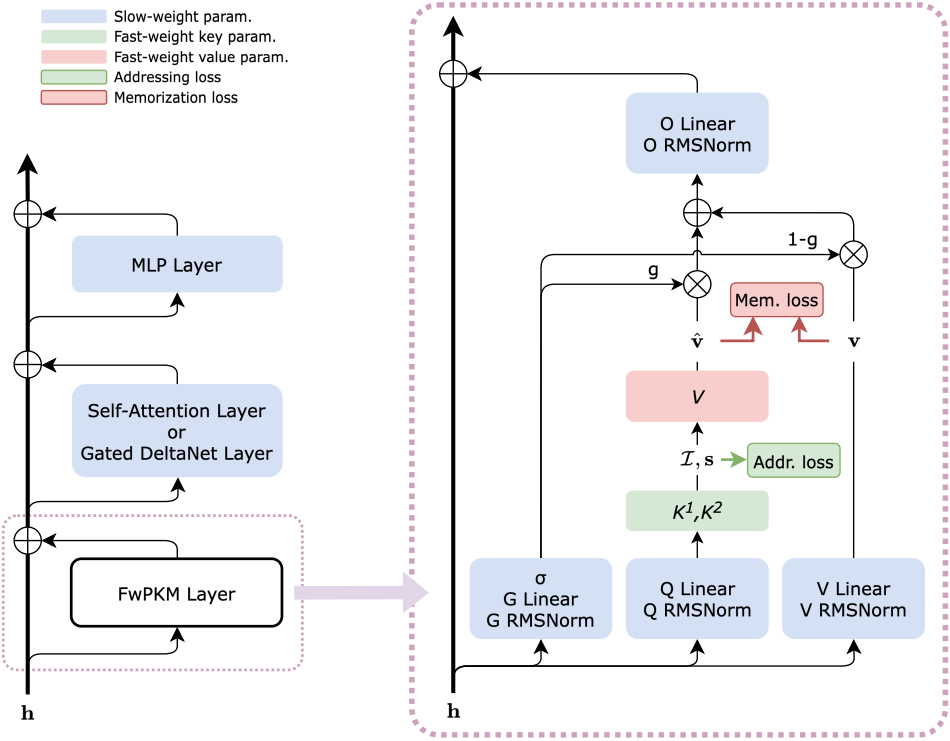}
        \caption{\footnotesize{\textbf{Architecture of \implname.} An FwPKM layer is placed before a token mixer (\textit{i.e.} a self-attention or a Gated DeltaNet layer). \cbox{slowBlue}{Slow-weight parameters} are updated at only training time, using a global language modeling loss. \cbox{fastRed}{Fast-weight value parameters} and \cbox{fastGreen}{key parameters} are updated at both training and inference time, by minimizing local \lossbox{borderRed}{fastRed}{memorization loss} and \lossbox{borderGreen}{fastGreen}{addressing loss}, respectively.}}
        \label{fig:fwpkm_arch}
    \end{center}
\end{figure}

Standard Product Key Memory (PKM) is a \textbf{slow-weight} sparse channel mixer, meaning that its parameters are learned by the global language modeling objective over a corpus and then remain frozen at inference.
As a result, PKM mainly stores \emph{semantic} knowledge but cannot rapidly incorporate \emph{episodic} information from the current input stream.
We propose \textbf{Fast-weight Product Key Memory} (\implname), which keeps PKM's \textbf{retrieval process unchanged}, but redesigns the \textbf{memorization process} by updating the memory matrices online.
Concretely, \implname~stores key-value associations from the input sequence by minimizing a local reconstruction objective over chunks, in the spirit of test-time training.

\subsection{Overview} \label{ssec:fwpkm_overview}
Figure~\ref{fig:fwpkm_arch} summarizes \implname.
Given hidden states $\mb{h}_{1:T}$\footnote{We abuse tensor subscripts to index both sequence and feature dimensions; $t$ indexes sequence position and $i,j$ index feature dimensions.}, we process them in chunks of size $C$.
Within each chunk, \implname:
(i) \textbf{constructs} query/value vectors and a scalar gate using slow-weight projections,
(ii) \textbf{retrieves} a sparse value prediction with PKM using the current fast weights,
and (iii) \textbf{writes} to the fast weights by optimizing local objectives over the chunk.
Concretely, we update the value matrix $V$ by minimizing a gated reconstruction loss $\mathcal{L}_{\text{mem}}$ (Section~\ref{ssec:mem_optim_main}), and update the key matrices $K^1,K^2$ by minimizing an auxiliary addressing loss $\mathcal{L}_{\text{addr}}$ that prevents slot collapse (Section~\ref{ssec:addr_optim}).
We apply these fast-weight updates after processing the chunk, so predictions within the chunk depend only on the fast weights from previous chunks, preserving causality.

\subsection{Forward Pass: Inputs and Retrieval} \label{ssec:fwpkm_forward}
\paragraph{Input construction.}
For each token $t$, a slow-weight network parameterized by $\phi$ constructs inputs to \implname:
\begin{align}
\mb{q}_t &= \mathtt{Linear}^{q}_\phi(\mathtt{RMSNorm}^{q}_\phi(\mb{h}_t)), \\
\mb{v}_t &= \mathtt{Linear}^{v}_\phi(\mathtt{RMSNorm}^{v}_\phi(\mb{h}_t)), \\
g_t &= \sigma\!\left(\mathtt{Linear}^{g}_\phi(\mathtt{RMSNorm}^{g}_\phi(\mb{h}_t))\right).
\end{align}
where $\mb{q}_t \in \mathbb{R}^{d_k}, \mb{v}_t \in \mathbb{R}^{d_v}, g_t \in (0,1)$ and $\sigma(\cdot)$ denotes the sigmoid function.

\paragraph{Sparse retrieval.}
\implname\ uses the same product-key sparse retrieval as PKM (Section~\ref{ssec:pkm}).
Given $\mb{q}_t$, PKM returns a sparse index set $\mathcal{I}_t$ and normalized weights $\{s'_{t,i}\}_{i\in\mathcal{I}_t}$, and predicts a value by a weighted sum of value rows:
\begin{align}
    \hat{\mb{v}}_{t} = \mathtt{PKM}(\mb{q}_t;\theta)
    = \sum_{i \in \mathcal{I}_t} s'_{t,i}\, V_i. \label{equ:fwpkm_retrieval}
\end{align}

\paragraph{Gated Residual Output.}
Next-token prediction does not always require episodic memory, so we interpolate the retrieved value with a residual value path:
\begin{align}
    \mb{o}_t = g_t \cdot \hat{\mb{v}}_{t} + (1-g_t)\cdot \mb{v}_t,
    \qquad
    \mb{o}'_t = \mathtt{Linear}^{o}_\phi(\mathtt{RMSNorm}^{o}_\phi(\mb{o}_t)). \label{equ:fwpkm_output}
\end{align}

\subsection{Backward Pass: Memorization by Learning to Reconstruct} \label{ssec:mem_optim_main}
\paragraph{Chunk-level local objective.}
Within each chunk, \implname\ updates fast weights $\theta$ to make retrieval reconstruct target values derived from the same stream.
Following TTT-style fast weights, we minimize a reconstruction loss over query-value pairs.
For a chunk of size $C$, define
\begin{align}
    \mathcal{L}_{\text{mem}}
    &= \sum_{t=1}^{C} \frac{1}{2}\, g_t \, \|\mb{v}_{t} - \hat{\mb{v}}_{t}\|_2^2. \label{equ:mem_loss}
\end{align}

\paragraph{Why MSE and the $\tfrac{1}{2}$ factor.}
The MSE gradient provides an explicit ``rewrite'' signal.
For a single prediction-target pair $(\hat{\mb{v}},\mb{v})$, the gradient with respect to the prediction is $\nabla_{\hat{\mb{v}}}\, \frac{1}{2}\|\mb{v}-\hat{\mb{v}}\|_2^2 = -(\mb{v}-\hat{\mb{v}})$. 
Thus, if we could update the prediction directly, a unit step would rewrite it to the target:
$\hat{\mb{v}} \leftarrow \hat{\mb{v}} - \nabla_{\hat{\mb{v}}} = \mb{v}$.
In \implname\ we do not update $\hat{\mb{v}}$ directly; instead we update the underlying value rows $V_i$ that produced $\hat{\mb{v}}$.
Nevertheless, the same residual term $(\mb{v}-\hat{\mb{v}})$ acts as the update signal, making the fast-weight update behave like explicit memory rewriting.

\paragraph{Gradient aggregation.}
Multiple tokens in a chunk may write to the same value row.
Let $N_i^{\text{read}}$ denote the number of times row $V_i$ is accessed in the chunk.
We aggregate per-row gradients by normalizing with $N_i^{\text{read}}$. This aggregation works as a consensus mechanism for competing memory writings.
\begin{align}
    \nabla^{\text{agg}}_{V_i}
    =
    \frac{1}{N^{\text{read}}_i} \nabla_{V_i} \mathcal{L}_{\text{mem}}. 
    \label{equ:V_update_agg}
\end{align}
For the memorization, we use $\nabla^{\text{agg}}_{V_i}$ to update the accessed value rows:
\begin{align}
V_i \leftarrow V_i - \nabla^{\text{agg}}_{V_i}.
\end{align}

\subsection{Backward Pass: Prevent Slot Collapsing via Addressing Optimization} \label{ssec:addr_optim}
Sparse memories can suffer from \textit{memory slot collapsing}, where only a small fraction of slots are used.
We therefore optimize an auxiliary addressing objective that encourages balanced slot usage on average across a chunk, without forcing each individual query to be uniform.

Let $\mb{s}^{\prime1}_t\in\mathbb{R}^{\sqrt{N}}$ and $\mb{s}^{\prime2}_t\in\mathbb{R}^{\sqrt{N}}$ be the normalized Top-$k$ score vectors over the two sub-key sets (unselected indices have score $0$).
Define marginal slot-usage distributions as:
\begin{align}
    \bar{\mb{p}}^{1} = \frac{1}{C}\sum_{t=1}^{C}\mb{s}^{\prime 1}_t,
    \qquad
    \bar{\mb{p}}^{2} = \frac{1}{C}\sum_{t=1}^{C}\mb{s}^{\prime 2}_t,
\end{align}
and minimize $\mathcal{L}_{\text{addr}}$ which is the negative entropy:
\begin{align}
    \mathcal{L}_{\text{addr}}
    &= -H(\bar{\mb{p}}^{1}) - H(\bar{\mb{p}}^{2}) \nonumber\\
    &= -\sum_{i=1}^{\sqrt{N}} \bar{p}^{1}_i \log \bar{p}^{1}_i \;-\;
       \sum_{i=1}^{\sqrt{N}} \bar{p}^{2}_i \log \bar{p}^{2}_i. \label{equ:addr_loss}
\end{align}
We update keys using the gradients of $\mathcal{L}_{\text{addr}}$:
\begin{align}
    K^1 \leftarrow K^1 - \nabla_{K^1}\mathcal{L}_{\text{addr}},
    \qquad
    K^2 \leftarrow K^2 - \nabla_{K^2}\mathcal{L}_{\text{addr}}. \label{equ:K_update}
\end{align}

\subsection{Practical Choices} \label{ssec:target_value_practical}
We found the following choices helpful for improving model performance and training stability. Detailed discussion and ablations are deferred to Appendix~\ref{app:fwpkm_impl} and Appendix~\ref{app:ablation}, respectively.
\begin{itemize}
    \item \textbf{Lookahead value targets.} During memorization, we pair $\mb{q}_t$ with $\mb{v}_{t+1}$. In this way, a query extracts information about the immediate future of similar queries in history, which aligns the local reconstruction task with the global next-token prediction task (Appendix~\ref{app:lookahead}).
    \item \textbf{IDW scoring.} Instead of scoring a key row by dot product $\mb{q}^{(m)\top} K^{(m)}_i, m\in\{1,2\}$ (Eq.~\ref{equ:pkm_score}), we use an inverse-distance weighting (IDW,~\citealt{IDW}) score that depends on Euclidean distance, which encourages keys to become \textit{clustering centroids} in representation space. We apply IDW by changing the scoring function from $\mb{q}^{\top} K_i$ to $- \log \big(\epsilon + \| \mb{q} - K_i \|_2^2 \big)$ (Appendix~\ref{app:idw}).
    \item \textbf{Target normalization.} We z-score normalize $\mb{v}_t$ along the feature dimension for stability (Appendix~\ref{app:target_norm}).
\end{itemize}

\section{Experiments} \label{sec:experiments}

We evaluate \implname~along four axes. First, we use perplexity on short- and long-context corpora to disentangle \textbf{semantic} (slow-weight) and \textbf{episodic} (fast-weight) memory roles. Next, we stress-test episodic retrieval with Needle-in-a-Haystack, including \textbf{iterative rereading} that leverages test-time optimization. Finally, we evaluate on realistic long-context QA tasks from LongBench and probe \textbf{online domain adaptation} on diverse Pile subsets.

%%%%%%%%%%%% Training setting %%%%%%%%%%%%%
\subsection{Training Setting} \label{ssec:training_setting}

\paragraph{\textbf{Model}}
We implement 12-layer language models that follow the QwenNext design principles~\citep{qwen3technicalreport,Qwen3Next80BThinking} but are scaled down to our setting (baseline models are $\sim$112M parameters), interleaving Gated DeltaNet (GDN,~\citealt{GDN}) and gated softmax attention layers~\citep{GatedAttn}. Our investigation uses the following baseline configurations.
\begin{itemize}[nosep]
    \item \texttt{GDN}: Gated DeltaNets at all 12 layers
    \item \texttt{GDN+SWA}: GDN interleaved with Sliding Window Attention (SWA, window size 512) at a 3:1 ratio
    \item \texttt{GDN+FA}: GDN interleaved with Full Attention (FA) at a 3:1 ratio
    \item \texttt{FA}: Full Attention at all layers
\end{itemize}
We introduce PKM and \implname~modules into these baselines at specific depths (layers 2, 6, and 10). A standard PKM layer replaces the original MLP, while an \implname~layer is inserted before the token mixer (Attention/GDN). Both PKM and \implname~have $512^2$ memory slots ($262{,}144$). PKM retrieves effectively Top-$128$ slots ($4$ heads and $32$ slots per head) while \implname~retrieves Top-$8$ slots ($1$ head and $8$ slots per head). We choose $8$ slots because it hits a balance between efficiency and performance. See ablation experiments in Appendix~\ref{app:ablation}.

We further include two types of baselines. For the first kind, we substitute the PKM architecture in \texttt{FwPKM@2,6,10} with a SwiGLU MLP~\citep{GLU} that maintains three fast-weight matrices and their biases for up-/gating-/down-projection. This variant is denoted as \texttt{FwMLP@2,6,10}. In addition, we use LaCT~\citep{TTTPlus} as the second baseline. LaCT is an improved TTT~\citep{TTT} model that uses a sliding window attention, a fast-weight SwiGLU MLP, and a slow-weight SwiGLU MLP in every layer. Its fast weights are updated using SGD+momentum with L2 weight normalization~\citep{WeightNormalization}.

\paragraph{\textbf{Data}}
To incentivize the learning of cross-chunk dependencies via \implname, we use 5B tokens from LongContext64~\citep{LC64}, a dataset consisting of long-context documents ($>64K$ tokens) sourced from RedPajama V2~\citep{RedPajama}. We supplement this with 5B tokens from Fineweb-Edu~\citep{Fineweb} to maintain high-quality language modeling capabilities. All models are trained with a sequence length of 4K tokens.

We refer readers to Appendix~\ref{app:exp_setting} for more details of the architectures and optimization.

%%%%%%%%%%%% PPL eval %%%%%%%%%%%%%
\subsection{PPL Evaluation} \label{ssec:ppl_eval}

\begin{figure}[t!]
    \begin{center}
        \includegraphics[width=0.98\textwidth]{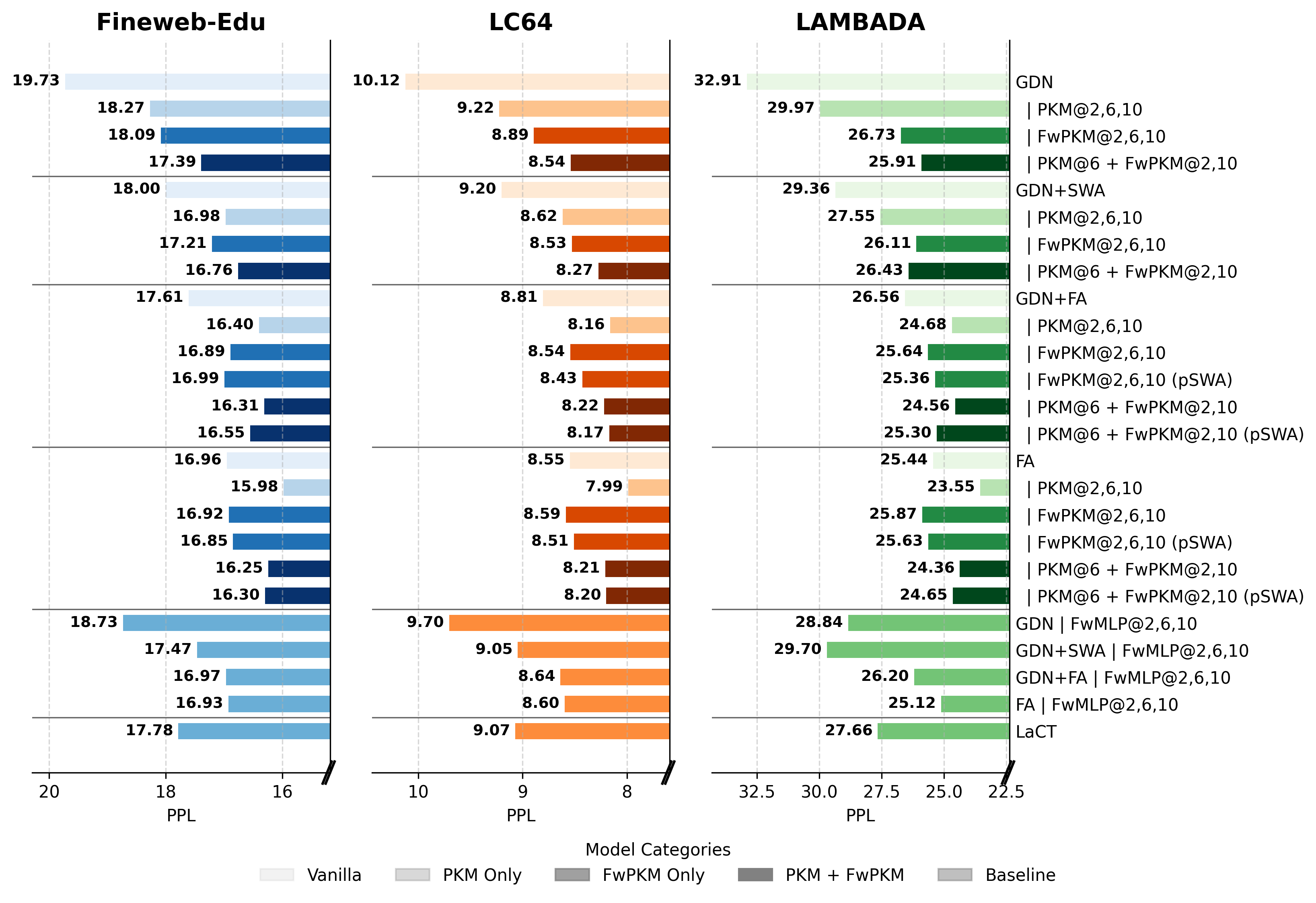}
    \end{center}
    \caption{\footnotesize{Perplexity on Fineweb-Edu, LC64, and LAMBADA. We use bar colors to help distinguish between models with different types of memory components.} \label{fig:eval_ppl}}
\end{figure}

\begin{figure}[t!]
    \begin{center}
        \includegraphics[width=0.95\textwidth]{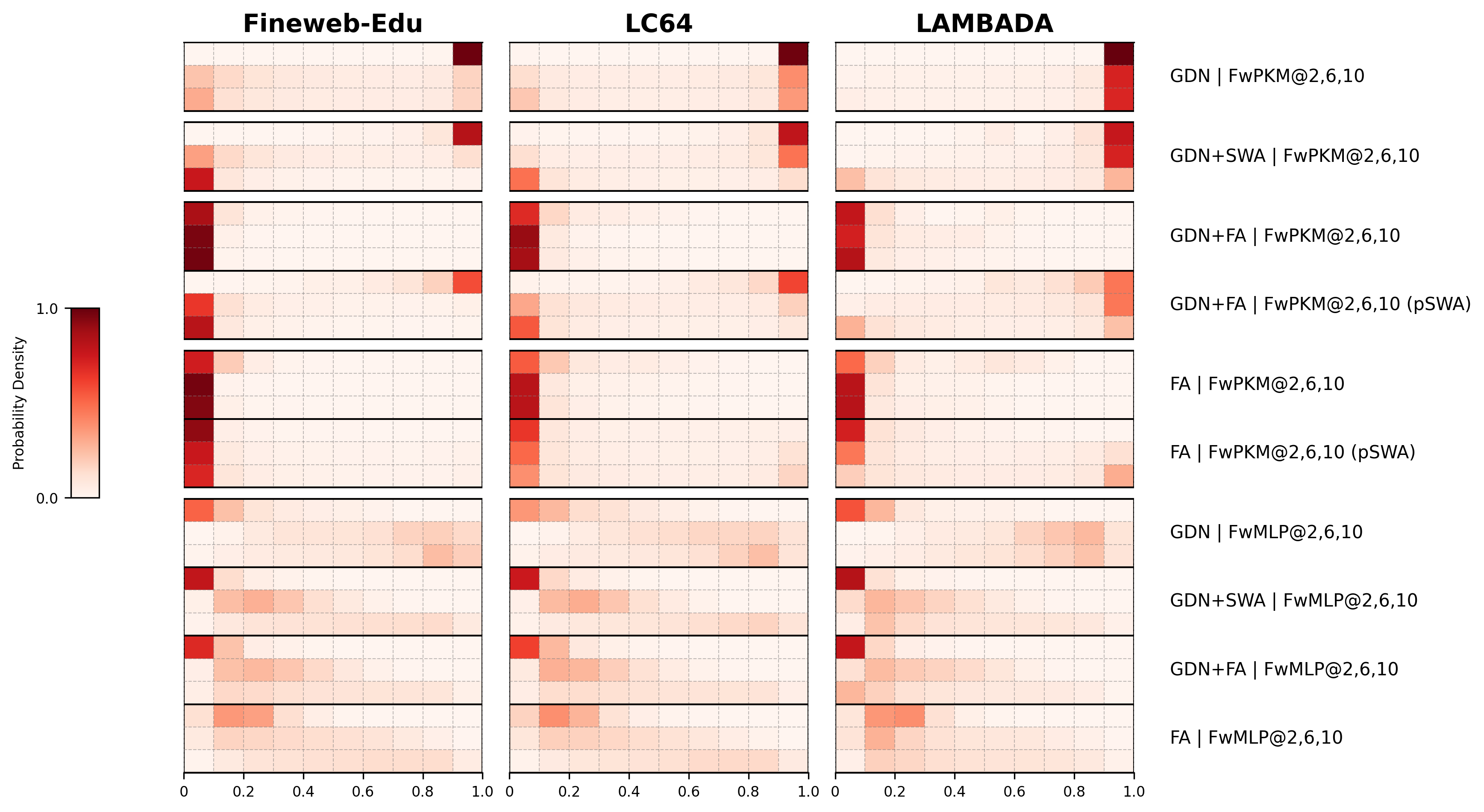}
    \end{center}
    \caption{\footnotesize{\implname~gating value distribution on Fineweb-Edu, LC64, and LAMBADA test sets. Each row represents one \implname~layer. The probability densities are categorized into 10 bins. A darker color suggests a higher probability density of a bin.}}
    \label{fig:eval_gate}
\end{figure}

We evaluate perplexity (PPL) on three distinct datasets to assess different memory capabilities: \textbf{Fineweb-Edu} (knowledge-intensive, short context), \textbf{LC64} (in-domain, long context), and \textbf{LAMBADA} (\citealt{LAMBADA}, out-of-domain, long context). Each evaluation dataset contains 8M tokens.

\paragraph{\textbf{Evaluation protocol}}
For PPL evaluation, we evaluate 4K-token sequences in their original order with a batch size of 1 so that segments are presented sequentially \emph{without resetting} fast weights between adjacent segments. This enables fast weights to capture dependencies across segment boundaries. Figure~\ref{fig:eval_ppl} shows perplexities for all models.

\paragraph{\textbf{Finding 1: \implname~and PKM serve distinct, complementary roles.}}
Baseline models with layouts \texttt{GDN} and \texttt{GDN+SWA} lack components for modeling long-range dependency, and their \implname~versions address the weakness. As shown in Figure~\ref{fig:eval_ppl}, \implname~significantly reduces PPL on long-context datasets (LC64, LAMBADA), confirming its role as an \textbf{episodic memory}. In contrast, standard PKM provides the largest gains on Fineweb-Edu, acting as a \textbf{semantic memory} for general knowledge. The combination of both modules yields the best performance across all metrics, suggesting they address orthogonal limitations in the baseline architectures. In Appendix~\ref{app:freeze_fw} and~\ref{app:ppl_diff_over_seq_boundary}, we further demonstrate that \implname~is likely to exploit cross-sequence dependency to reduce PPL on long-context datasets.

\paragraph{\textbf{Finding 2: \implname~competes with Full Attention.}}
When \implname~is added to baselines with unrestricted Full Attention (\texttt{FA} and \texttt{GDN+FA}), PPL improvements are marginal. Analysis of gating values in Figure~\ref{fig:eval_gate} reveals that these models learn to ignore \implname, with gating weights clustering near zero.

\paragraph{\textbf{Finding 3: Restricting full attention facilitates \implname~use.}}
To mitigate the above issue, we restrict full attention's long-range perception at \emph{training time} by imposing a sliding attention window of length 512 on all full attention layers with probability 0.9. Denoted by a suffix \texttt{pSWA}, these models learn to use \implname~more, as suggested by increased high-end mass in the gating distributions in Figure~\ref{fig:eval_gate}. While \texttt{pSWA} has minimal impact on PPL, we will see a more significant difference in the following experiments.

%%%%%%%%%%%% NIAH eval %%%%%%%%%%%%%
\subsection{Needle-in-a-Haystack Evaluation} \label{ssec:niah_eval}

We conduct Needle-in-a-Haystack (NIAH,~\citealt{NIAH,LandmarkAttention}) evaluation to further verify \implname's functionality as episodic memory. In all NIAH experiments, we update fast weights after reading the full haystack (and after each \textit{rereading} iteration, see below); we do not perform additional updates while generating the answer.

In the basic setting, we construct 500 NIAH samples from the LAMBADA dataset. Each sample contains a haystack -- a 4K-length sequence from the LAMBADA dataset with 5 needles inserted at random positions, where each needle contains a unique 4-character key and a 6-digit value -- and a question that requests the value of a specific key. To test \implname's large memory storage, we additionally construct test sets of 8K, 32K, and 128K context lengths.

\textbf{Iterative Memorization ($n$-iter NIAH).}
A unique feature of \implname~is its ability to improve memory fidelity by re-processing the same input. We define a \textbf{$n$-iter NIAH} setting, where the model forwards the same haystack context $n$ times before answering. \implname's process chunk size $C$ is accordingly changed to the length of each haystack context such that it updates its memory once after reading an entire haystack. The basic setting is $1$-iter NIAH.

\begin{figure}[t!]
    \begin{center}
        \includegraphics[width=1.0\textwidth]{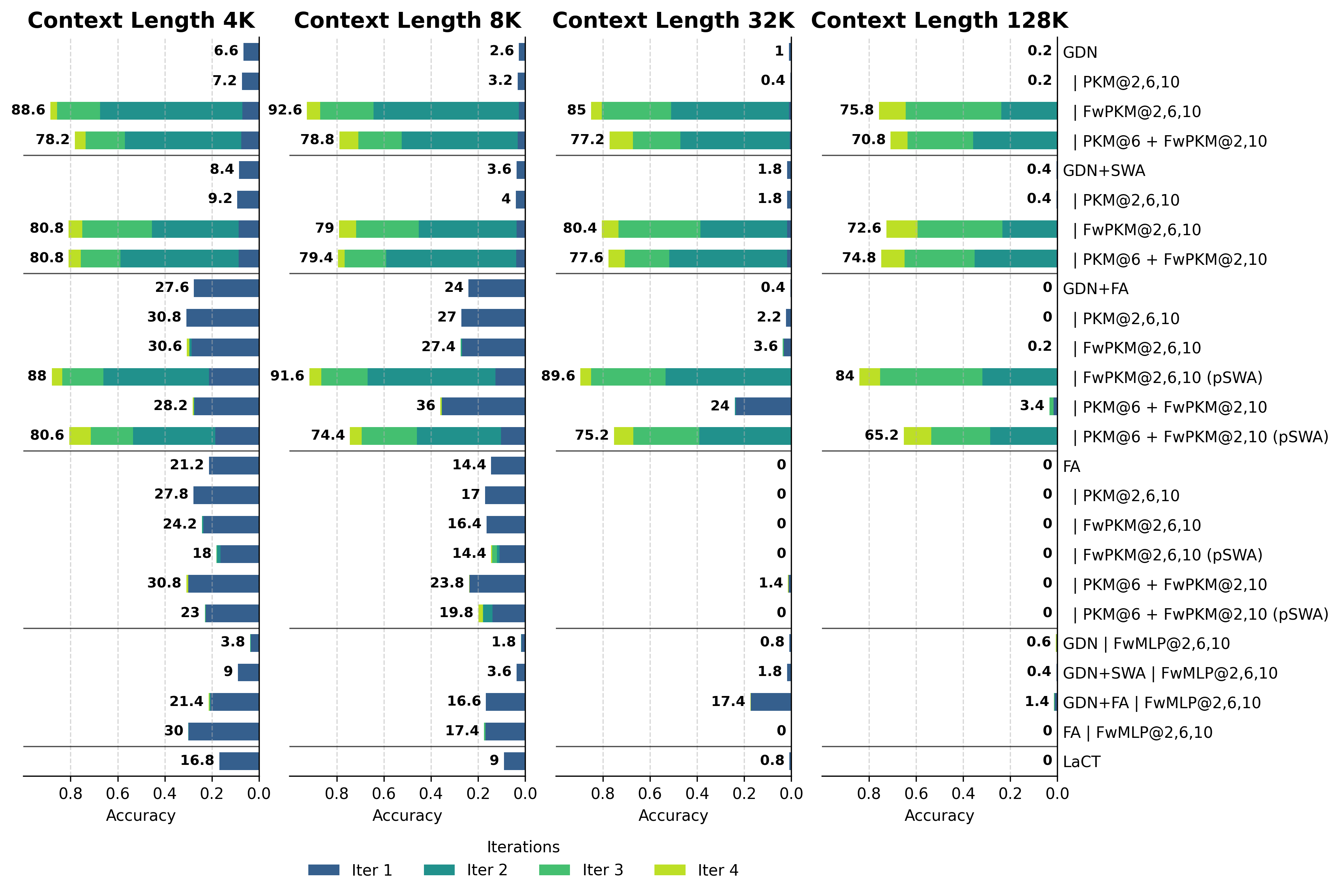}
    \end{center}
    \vspace{-5pt}
    \caption{\footnotesize{Stacked bar plots for NIAH accuracy results on 4K-/8K-/32K-/128K-length test sets. Each stacked bar shows the accuracies of $\{1,2,3,4\}$-iter NIAH evaluations.} \label{fig:eval_niah}}
\end{figure}

Accuracy results of $n$-iter NIAH ($n \in \{1,2,3,4\}$) on the four test sets of 4K-128K context lengths are shown in Figure~\ref{fig:eval_niah}. The accuracies of all iterations are drawn as a stacked bar.

\paragraph{\textbf{Finding 4: Iterative reading boosts retrieval accuracy.}}
For \texttt{GDN} and \texttt{GDN+SWA} layouts, a single pass ($1$-iter) is often insufficient for perfect retrieval. However, a second pass ($2$-iter) yields a massive boost in accuracy (jumping from $<10\%$ to $>50\%$ in many cases). More passes further improve the accuracy to $>70\%$. This confirms that \implname~effectively exploits test-time training to consolidate episodic memories, surpassing softmax attention. In addition, effective iterative memorization (in Figure~\ref{fig:eval_niah}) correlates with high gating values (in Figure~\ref{fig:eval_gate}).

\paragraph{\textbf{Finding 5: \implname~generalizes to 128K contexts.}}
Despite being trained on only 4K-token sequences, \implname~generalizes effectively to 128K tokens. While \texttt{FA} baselines degrade rapidly on context lengths unseen during training, \implname~maintains robust retrieval performance.

%%%%%%%%%%%% Longbench eval %%%%%%%%%%%%%

\begin{figure}[!]
    \begin{center}
        \includegraphics[width=0.97\textwidth]{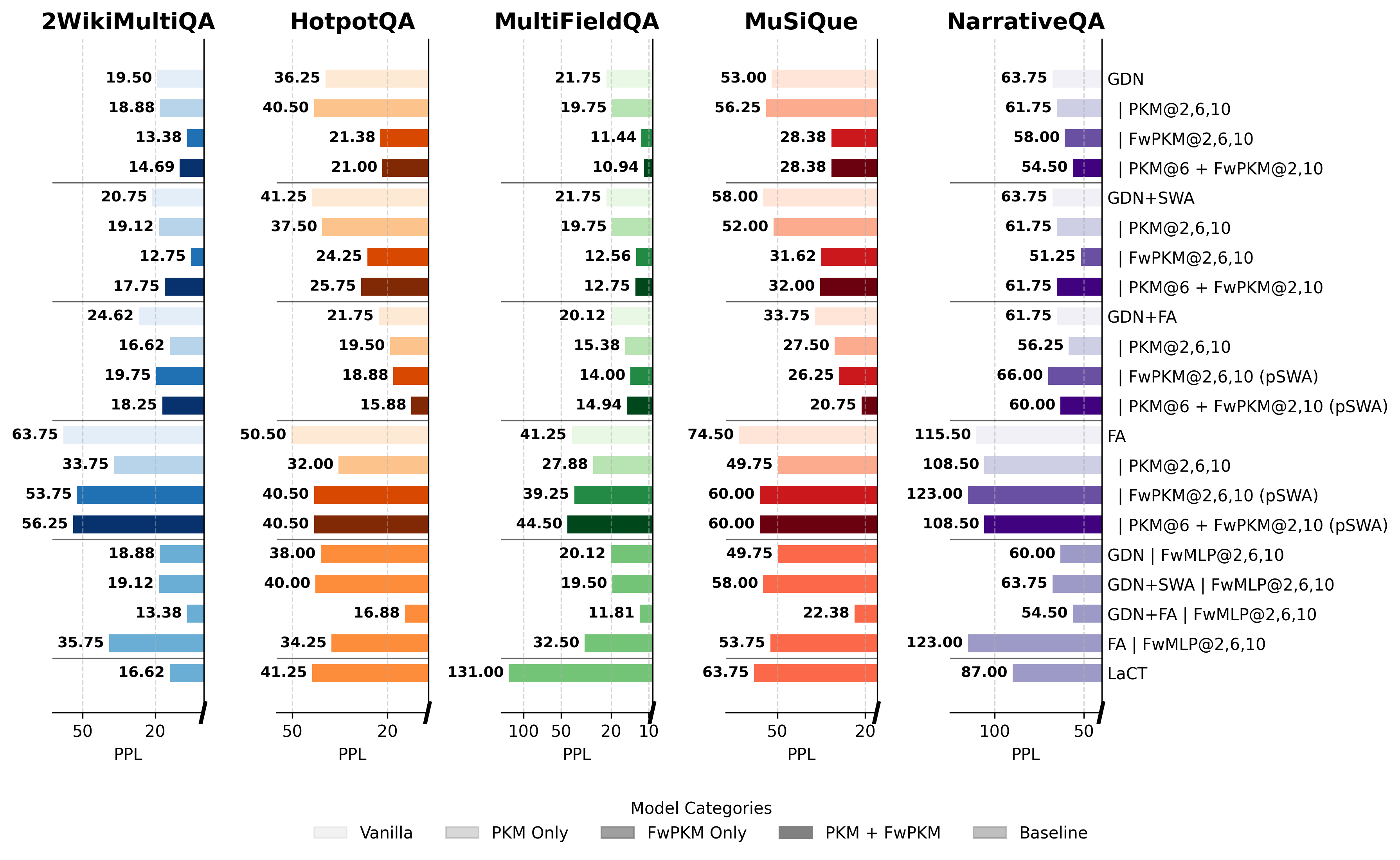}
    \end{center}
    \vspace{-5pt}
    \caption{\footnotesize{Perplexity on five Longbench tasks. We use bar colors to help distinguish between models with different types of memory components.} \label{fig:eval_longbench}}
\end{figure}

\subsection{Longbench Evaluation} \label{ssec:longbench_eval}

In addition to the synthetic NIAH task, we assess long-context modeling on realistic data from Longbench~\citep{Longbench}. Longbench contains 21 tasks whose average length ranges from 5K to 15K whitespace-separated words. The tasks are mostly question answering and summarization, but their generative nature makes it difficult for pretrained-only (non instruction-tuned) models to produce well-calibrated free-form answers. Instead of computing official metrics such as F1 score, we use perplexity for evaluation.

We select \textit{2WikiMultihopQA}, \textit{HotpotQA}, \textit{MultiFieldQA-en}, \textit{MuSiQue}, and \textit{NarrativeQA}, resulting in 950 examples with an average length of 17421 tokens and a maximum length of 81616 tokens. We only compute perplexity on answer tokens. For examples with multiple answer references, we use the first reference for perplexity computation. We update fast weights once after reading the entire context (i.e., no updates during answer generation), mirroring the $1$-iter NIAH protocol. We omit the data for \texttt{LaCT} because it performs poorly on the Longbench data (PPL $>100$).

\paragraph{\textbf{Finding 6: \implname~reduces answer-token perplexity on realistic long-context tasks.}}
As shown in Figure~\ref{fig:eval_longbench}, \implname~effectively reduces perplexity on these long-context tasks for \texttt{GDN}, \texttt{GDN+SWA}, and \texttt{GDN+FA} models where \implname~is actively utilized.

%%%%%%%%%%%% Continual learning eval %%%%%%%%%%%%%
\subsection{Continual Learning Evaluation} \label{ssec:pile_domain_eval}

\begin{figure}[!]
    \begin{center}
        \includegraphics[width=0.8\textwidth]{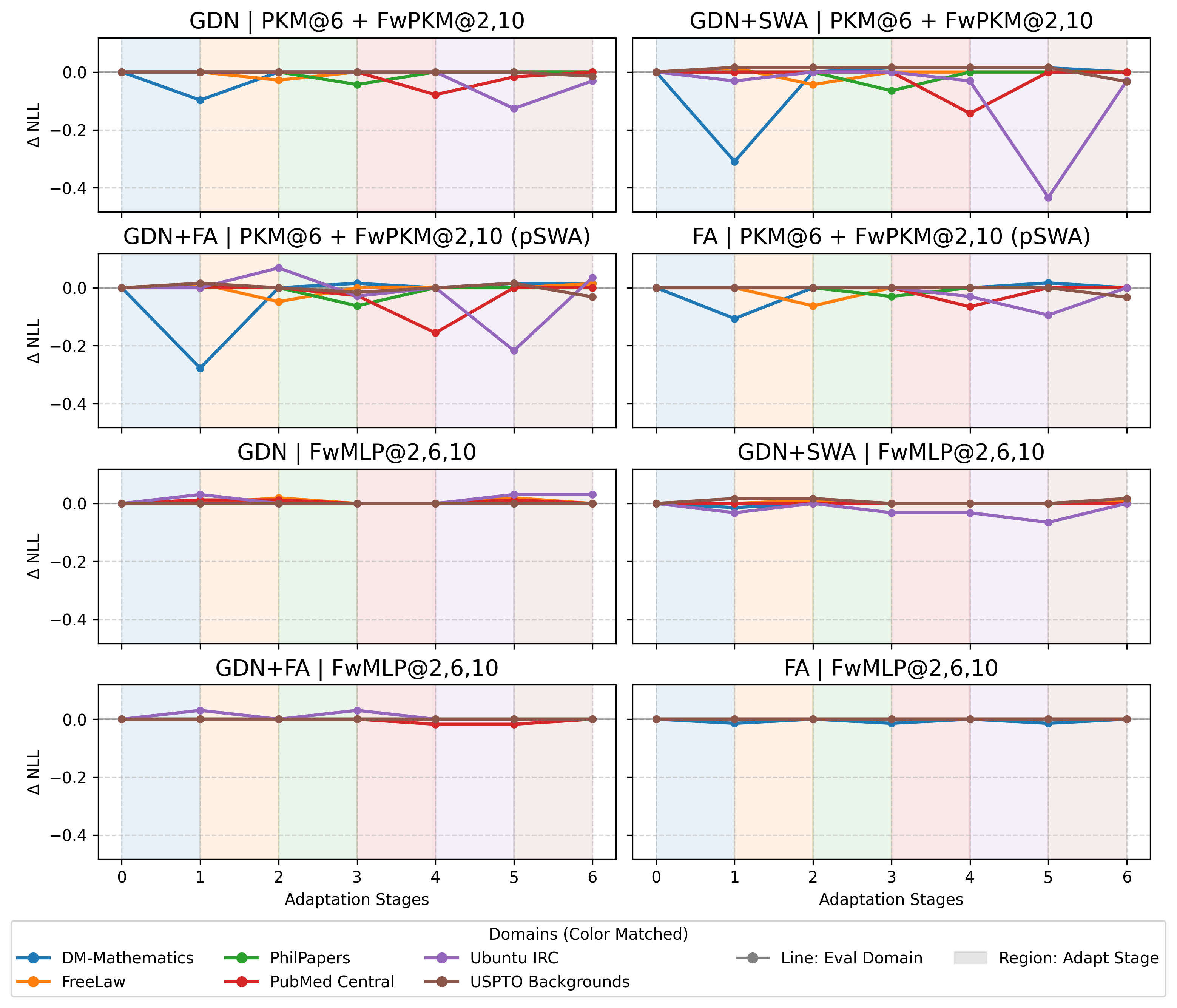}
    \end{center}
    \vspace{-5pt}
    \caption{\footnotesize{Change of negative log-likelihood ($\Delta$NLL) on six Pile domain datasets before and after 6 adaptation stages of test-time learning on each domain dataset.} \label{fig:eval_pile_domain}}
\end{figure}

Beyond long-context modeling, we study whether the large-scale fast weights in \implname~adapt to new data domains. We devise an \textit{adapt-and-test} experiment. From the Pile corpus~\citep{Pile}, we choose six subsets \textit{DM Mathematics}, \textit{FreeLaw}, \textit{PhilPapers}, \textit{PubMed Central}, \textit{Ubuntu IRC}, and \textit{USPTO Backgrounds} to represent different professional domains. For each domain, we sample a 4M-token \textit{adapt set} and a 4M-token \textit{eval set} without overlap.

We run 6 adaptation stages sequentially (one per domain's \textit{adapt set}), during which fast weights are updated while slow weights are frozen. After each adaptation stage, we compute the model's average negative log-likelihood (NLL) on all 6 \textit{eval sets} and report changes relative to the initial NLL before the first stage. In Figure~\ref{fig:eval_pile_domain}, domains correspond to colors, used both for the background shading of each adaptation stage and for the lines showing $\Delta\text{NLL}$.

\paragraph{\textbf{Finding 7: \implname~adapts quickly but struggles with retention across stages.}}
As expected, after each adaptation stage, the NLL of the corresponding domain decreases as the fast weights store related domain knowledge. However, these gains often do not persist across subsequent stages, suggesting that previously stored fast-weight knowledge is flushed and replaced by newer domains. This motivates developing a memory retention mechanism to realize long-term continual learning in future work.

\section{Interpretability Analyses}  \label{sec:interpret}

A key advantage of \implname~over black-box architectures is its inherent interpretability. Because memory slots are explicitly written and read, we can trace retrieved information back to specific input tokens.

\subsection{Case Study: Probing Memory in NIAH} \label{ssec:prob_niah}

\begin{figure*}
    \begin{center}
        \includegraphics[width=0.97\textwidth]{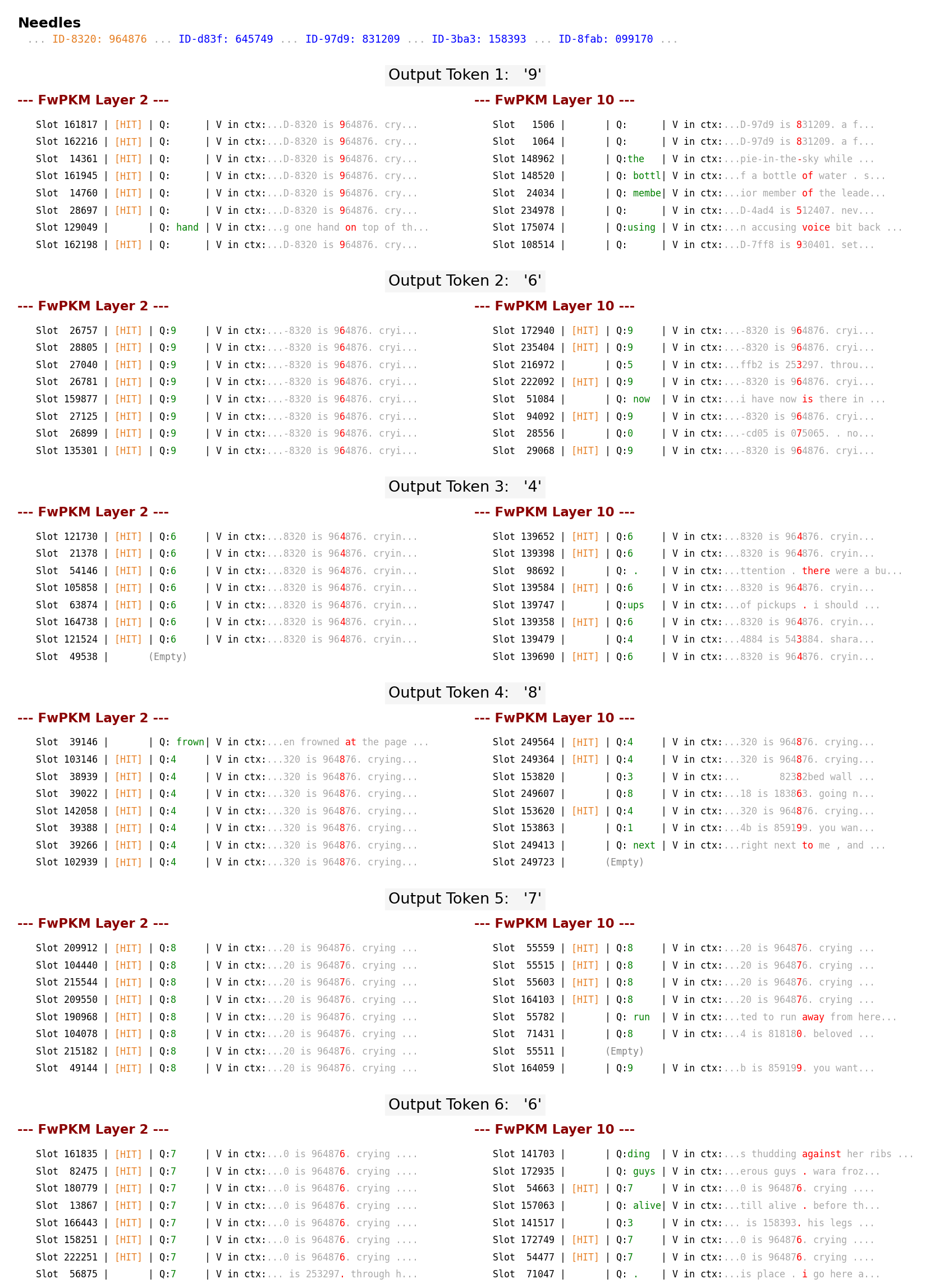}
    \end{center}
    \caption{\footnotesize{An example of \implname~slot access of \texttt{GDN+SWA|PKM@6+FwPKM@2,10} during generating an NIAH-4K answer. The model memorizes the haystack for 3 extra iterations, \textit{i.e.} $4$-iter NIAH. See Section~\ref{ssec:prob_niah} for the visual elements' definitions.} \label{fig:mem_inspection}}
\end{figure*}

We investigate the internal retrieval mechanism by visualizing the memory access patterns of the \texttt{GDN+SWA|PKM@6+FwPKM@2,10} model during a 4-iter Needle-in-a-Haystack (NIAH) task with a 4K context. Specifically, we trace the model's internal state as it generates the 6-digit passkey answer. 

Figure~\ref{fig:mem_inspection} displays the Top-$8$ retrieved memory slots for both a lower layer (Layer 2) and a higher layer (Layer 10) at each generation step. For every retrieved slot, we decode and display the content stored during the latest memorization step: the \textcolor{Green}{query token} (the key used for addressing), the \textcolor{Red}{target value token} (the predicted next token), and the \textcolor{Gray}{surrounding context} (the text window surrounding the target in the original haystack). To facilitate analysis, we mark slots with a \textcolor{Orange}{\texttt{[HIT]}} label if they explicitly store the ground-truth next token required for the current generation step.

\paragraph{\textbf{High-precision retrieval}} As shown in Figure~\ref{fig:mem_inspection}, the majority of retrieved slots contain the correct target tokens associated with the query needle. The model successfully surfaces the specific needle (ID ``8320'' at depth $89.66\%$) from thousands of tokens, populating the top slots with the correct 6-digit value sequence (``964876'').

\paragraph{\textbf{Robust aggregation}} Despite individual slot errors, the model successfully aggregates information across the $16$ slots from the two \implname~layers to generate the correct 6-digit value. This demonstrates that \implname~functions as a robust distributed storage system, where consensus across multiple slots compensates for noise in individual retrievals.

\subsection{Case Study: Selective Gating}

\begin{figure*}[t!]
    \begin{center}
        \includegraphics[width=0.95\textwidth]{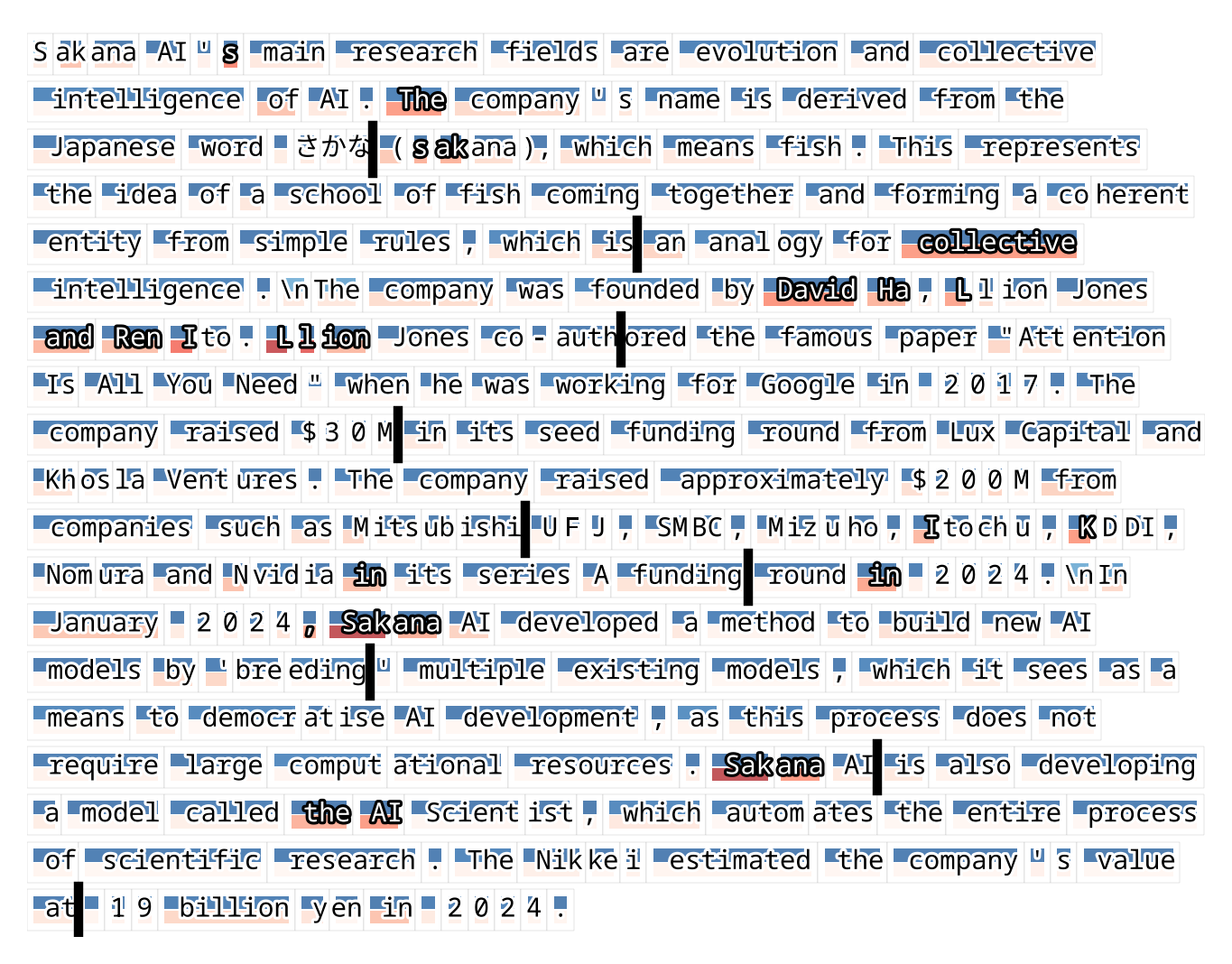}
    \end{center}
    \caption{\footnotesize{\texttt{GDN+SWA|PKM@6+FwPKM@2,10}'s \implname~gating values on tokens from the Wikipedia article for Sakana AI. The blue color in the background denotes the gating intensity of the \implname~at layer 2 and the red color denotes the \implname~at layer 10. A darker color represents higher intensity. Black vertical lines show positions where \implname's fast weights are updated, which is every 32 tokens as we specified for this visualization example.}\label{fig:gating_example}}
\end{figure*}

We further examine the ``gating values'' $g_t$ to understand when the model chooses to rely on episodic memory versus its static slow weights. To visualize this, we feed the Wikipedia article for ``Sakana AI''\footnote{\url{https://en.wikipedia.org/wiki/Sakana_AI}}—a topic unseen during the model's pre-training—into the network. We adjust the update chunk size from 512 to 32 to capture fine-grained local adaptivity.

Figure~\ref{fig:gating_example} presents a heatmap of the gating scalar $g_t$ with input text in the foreground. Higher intensity (represented by darker bars) indicates a value of $g_t$ closer to $1$, signifying a strong reliance on the fast-weight episodic memory. Conversely, lower intensity indicates reliance on the pre-trained slow weights.

\paragraph{\textbf{Layer specialization}} We observe distinct behaviors across model depths. The lower-layer \implname~tends to maintain high gating values across almost all tokens, acting as a general-purpose buffer that indiscriminately caches recent history. In contrast, the higher-layer \implname~exhibits highly selective activation.

\paragraph{\textbf{Novelty detection}} As seen in Figure~\ref{fig:gating_example} (layer 10), gating values spike specifically for tokens related to rare named entities and proper nouns (e.g., ``\textit{Sakana AI}'', ``\textit{David Ha}'', ``\textit{Llion Jones}'', and ``\textit{Ren Ito}''). This indicates that the model effectively distinguishes between general linguistic patterns (processed by slow weights) and novel, context-specific entities (processed by fast weights).

% \FloatBarrier

\section{Cost Analyses} \label{sec:cost}

We evaluate the scalability and efficiency of \implname~by profiling model size and computational costs, summarized in Table~\ref{tab:model_comparison}. Our analysis focuses on three key metrics:
\begin{itemize}[nosep]
    \item \textbf{Active Parameters}: The subset of parameters actually used during a forward pass, highlighting the architectural sparsity.
    \item \textbf{FLOPs} (Floating Point Operations): The theoretical computational volume required for each run.
    \item \textbf{FLOPS} (FLOPs per Second): The empirical hardware utilization efficiency, derived from actual running time.
\end{itemize}

Profiling was conducted by running 100 training steps on a single H100 GPU with a global batch size of 8, using the configuration detailed in Section~\ref{ssec:training_setting} and Appendix~\ref{app:exp_setting}.

\paragraph{\textbf{Sparsity and Computational Volume}}
Despite a substantial increase in total parameter count (approx. $5\times$ larger than baselines), the \textbf{Active Parameter} count of \implname~models remains comparable to the dense baselines (\textit{e.g.}, \texttt{GDN}: 112M vs. \texttt{FwPKM}: 116M). Consequently, the theoretical computational cost (\textbf{FLOPs}) is remarkably efficient; in some cases, the sparsity of PKM/\implname~makes them even less FLOPs-intensive than standard dense MLP layers.

\paragraph{\textbf{Hardware Utilization and Kernels}}
While theoretical costs are low, the wall-clock speed (\textbf{Samples per sec.}) and hardware utilization (\textbf{FLOPS}) reveal an implementation gap. \implname~components currently run slower than their dense counterparts. This is primarily due to the maturity difference in kernel optimization: softmax attention and GDN benefit from highly optimized kernels like FlashAttention~\citep{FlashAttention,FlashAttention2} and FlashLinearAttention~\citep{FLA}. In contrast, our sparse operations rely on less specialized implementations. Developing dedicated, hardware-aware kernels for sparse fast-weight updates is a critical future direction to bridge this gap and facilitate broader adoption.

\begin{table}[t!]
\caption{\footnotesize{Comparison of model size and computation cost. Active parameter counts reflect the sparse layer specifications: PKM utilizes 128 active rows, while \implname~utilizes 8 active rows per token.} \label{tab:model_comparison}}
\begin{center}
\begin{small}
\begin{tabular}{lrrrrr}
\toprule
\multirow{2}{*}{\textsc{Model}} & \multicolumn{2}{c}{Parameters (M)} & \multirow{2}{*}{FLOPs (T)} & \multirow{2}{*}{FLOPS (T/sec.)} & \multirow{2}{*}{Samples/sec.} \\
\cmidrule(lr){2-3}
 & Total & Active & & & \\
\midrule

\texttt{GDN} & 112.11 & 112.11 & 22.04 & 150.00 & 54.44 \\
\texttt{  | PKM@2,6,10} & 509.08 & 106.62 & 20.92 & 112.19 & 42.90 \\
\texttt{  | FwPKM@2,6,10} & 519.12 & 116.48 & 22.74 & 54.12 & 19.04 \\
\texttt{  | PKM@6 + FwPKM@2,10} & 515.77 & 113.19 & 22.14 & 58.85 & 21.27 \\
\midrule
\texttt{GDN+SWA} & 112.65 & 112.65 & 22.15 & 151.11 & 54.58 \\
\texttt{  | PKM@2,6,10} & 509.61 & 107.15 & 21.03 & 109.43 & 41.63 \\
\texttt{  | FwPKM@2,6,10} & 519.66 & 117.02 & 22.85 & 51.70 & 18.10 \\
\texttt{  | PKM@6 + FwPKM@2,10} & 516.31 & 113.73 & 22.24 & 58.55 & 21.06 \\
\midrule
\texttt{GDN+FA} & 112.65 & 112.65 & 22.16 & 147.71 & 53.33 \\
\texttt{  | PKM@2,6,10} & 509.61 & 107.15 & 21.04 & 102.76 & 39.07 \\
\texttt{  | FwPKM@2,6,10} & 519.66 & 117.02 & 22.86 & 51.91 & 18.16 \\
\texttt{  | PKM@6 + FwPKM@2,10} & 516.31 & 113.73 & 22.25 & 57.69 & 20.74 \\
\midrule
\texttt{FA} & 114.25 & 114.25 & 22.51 & 158.75 & 56.42 \\
\texttt{  | PKM@2,6,10} & 511.22 & 108.76 & 21.39 & 106.40 & 39.79 \\
\texttt{  | FwPKM@2,6,10} & 521.26 & 118.62 & 23.21 & 54.34 & 18.73 \\
\texttt{  | PKM@6 + FwPKM@2,10} & 517.91 & 115.33 & 22.60 & 61.80 & 21.87 \\
\midrule
\texttt{GDN | FwMLP@2,6,10} & 118.05 & 118.05 & 23.21 & 134.04 & 46.21 \\
\texttt{GDN+SWA | FwMLP@2,6,10} & 118.58 & 118.58 & 23.31 & 128.40 & 44.06 \\
\texttt{GDN+FA | FwMLP@2,6,10} & 118.58 & 118.58 & 23.32 & 97.18 & 33.33 \\
\texttt{FA | FwMLP@2,6,10} & 120.19 & 120.19 & 23.67 & 118.85 & 40.16 \\
\midrule
\texttt{LaCT} & 112.70 & 112.70 & 24.44 & 92.38 & 30.24 \\

\bottomrule
\end{tabular}
\end{small}
\end{center}
\end{table}
\section{Related Work}

\textbf{Softmax attention and linear variants}
Standard softmax attention has been foundational to the success of Transformers~\citep{Transformer}, fundamentally acting as a powerful form of associative memory~\citep{TransformerAssocMem}. However, its quadratic complexity limits its application to extremely long sequences. To address this, various efficient architectures have been proposed. Linear attention~\citep{LinearAttention} reduces complexity to linear time by changing the order of association. This direction includes Recurrent Neural Network (RNN) and State Space Model (SSM) evolutions such as Mamba~\citep{Mamba} and Mamba2~\citep{Mamba2}, as well as specific variants like DeltaNet~\citep{DeltaNetLSTM, ParallelDeltaNet}, Gated DeltaNet~\citep{GDN}, and RWKV7~\citep{RWKV7}. Other approaches like Memory Mosaics~\citep{MemoryMosaics, MemoryMosaicsPlus} improve softmax attention by techniques such as time-dimension smoothing and hierarchical memory design.

\textbf{Fast weights and test-time training}
The concept of ``fast weights'' offers a powerful lens for unifying sequence modeling. Rooted in early work by \citet{FW-Schmidhuber} and \citet{Ba-FW}, this framework views linear transformers as fast weight programmers~\citep{FWLinearAttention}. Recently, this paradigm has been revitalized by Test-Time Training (TTT)~\citep{TTT, TTTPlus} and Titans~\citep{Titans}, which explicitly update parameters during inference using gradient descent, enabling the model to memorize the current context. Theoretical frameworks like MIRAS~\citep{MIRAS} and Test-Time Regression~\citep{TTRegression} unified various sequence models under the umbrella of test-time optimization and associative memory. These frameworks pointed out several directions for new model designs, namely memory architecture, memorization rule, memory retention rule, and optimizer. Our proposal of \implname~contributes a novel memory architecture and a specific memorization rule accommodated for its structural sparsity.

\textbf{Hybrid architectures}
Recognizing the complementary strengths of different sequence models, recent work has increasingly focused on hybrid architectures that combine the high-fidelity retrieval of quadratic attention with the efficiency of linear or recurrent layers. This includes hybrid models trained from scratch~\citep{HybridFAandDeltaNet}, as well as large-scale Hybrid LLMs such as Samba~\citep{Samba} and KimiLinear~\citep{KimiLinear}. QwenNext~\citep{qwen3technicalreport,Qwen3Next80BThinking} also employs this interleaved design. Furthermore, approaches like Artificial Hippocampus Networks (AHN,~\citealt{AHN}) explore fine-tuning techniques to integrate these distinct memory systems effectively. Experiments in Section~\ref{sec:experiments} demonstrated interactions between memories of different characteristics. The combination of linear attention (\textit{i.e.} GDN), softmax attention, slow-weight sparse memory (\textit{i.e.} PKM), and fast-weight sparse memory (\textit{i.e.} \implname) implements a versatile memory system that excels at various tasks.

\textbf{Memory models}
Beyond implicit knowledge in weights, explicit memory modules have been explored to enhance storage capacity. Early works include Memory Networks~\citep{MemNet-JW}, while recent studies suggest even simple MLPs can function as memory~\citep{TwoLayerMLPMem}. To scale up capacity without prohibitive costs, sparse access mechanisms are essential. Product Key Memory (PKM,~\citealt{PKM, PKMPlus}) and PEER~\citep{PEER} utilize sparsity to access massive memory banks efficiently. Ultra Sparse Memory~\citep{UltraMem,UltraMemV2} is a line of work that extends the PKM architecture with more expressive keys and other improvements. Our work builds upon the PKM structure but transitions it from a static ``slow'' memory to a dynamic ``fast'' memory.

\textbf{Continual and episodic learning}
Finally, the ability to update memory parameters allows for continual learning and adaptation. \citet{PKMFinetune} showed that parameter-efficient fine-tuning of PKM effectively mitigates catastrophic forgetting. It demonstrated one dimension for continual learning -- optimizing sparse slow weights via self-supervised learning to update semantic memory. \implname~opens up the possibility for a different dimension of updating episodic memory by learning fast weights in an online learning manner. Nested Learning~\citep{NestedLearning} and TNT~\citep{TNT} explored the direction of stacking multiple fast weight layers (\textit{e.g.} Titans) and apply memory updates at varying frequencies. This ``nested'' memory structure enables gradually digesting changes in faster weights into slower weights and exhibits strong performance. \implname~maintains a huge memory bank and is updated at a low frequency to amortize optimization cost. It is a promising direction to design a hybrid memory system of varying-size \implname\textit{s} and other memory components such as Titans with different optimization strategies as in Nested Learning~\citep{NestedLearning}.

\section{Conclusion}

In this work, we introduced \textbf{Fast-weight Product Key Memory (\implname)}, a novel architecture that resolves the tension between memory capacity and computational efficiency in sequence modeling. By unifying the large-scale sparse storage of Product Key Memory (PKM) with the rapid adaptability of Test-Time Training (TTT), we transform PKM from a static retrieval module into a dynamic, context-responsive component. \implname~performs \textit{Sparse TTT} -- updating key-value parameters online via sparse gradient descent -- and enables models to efficiently memorize vast amounts of transient information and retrieve it thousands of tokens later.

Empirically, \implname~demonstrates remarkable robustness. Models trained on short 4K-token sequences generalize effectively to 128K-token contexts (\textit{i.e.} a $32\times$ length extrapolation) without fine-tuning. Furthermore, we show that iterative reading exploits the TTT nature of \implname, boosting retrieval accuracy from $<10\%$ to $>70\%$ in difficult Needle-in-a-Haystack scenarios. Our ablation study and interpretability analyses confirm that \implname~functions as a true episodic memory, selectively activating for novel entities and complementing the semantic knowledge stored in standard slow weights.

Despite these successes, challenges remain. The introduction of online updates introduces additional computational overhead, currently limited by the lack of optimized kernels for sparse updates. This motivates future systems work on hardware-aware sparse kernels to close the speed gap with dense baselines. On the modeling side, our continual learning experiments reveal that while \implname~adapts rapidly, it struggles with long-term retention across domain shifts. Future research should focus on advanced retention mechanisms and memory consolidation strategies to prevent catastrophic forgetting of fast weights. Overall, \implname~represents a promising step toward language models with versatile and mutually complementary memory systems.
\section*{Acknowledgments}

We would like to thank Kai Arulkumaran, Luke Darlow, Stefania Druga, and the entire Sakana AI team for their insightful discussions and support throughout this project. We also extend our gratitude to Huayang Li, Qianying Liu, Richard Sproat, Yujin Tang, Qi Sun, and Stefano Peluchetti for their valuable feedback, which significantly strengthened the quality and clarity of this manuscript.

\newpage
\bibliography{main}
\bibliographystyle{plainnat}

\clearpage
\appendix

\section{\implname~Implementation Details} \label{app:fwpkm_impl}

\subsection{Lookahead Value Targets} \label{app:lookahead}
By default, \implname\ pairs each query $\mb{q}_t$ with the value computed from the same token, $\mb{v}_t$, in the memorization loss (Eq.~\ref{equ:mem_loss}).
However, next-token prediction benefits more directly from information about \textit{future} tokens.
We therefore use a \textbf{lookahead} construction that pairs $\mb{q}_t$ with the next token's value $\mb{v}_{t+1}$ when applying chunk-level updates.

Concretely, we keep the retrieval definition $\hat{\mb{v}}_{t}=\mathtt{PKM}(\mb{q}_t;\theta)$ unchanged, and only shift the target value in the memorization objective. We use gradients from this new objective to update fast-weight value matrix $V$.
\begin{align}
    \mathcal{L}_{\text{mem}}^{\text{LA}}
    &= \sum_{t=1}^{C-1} \frac{1}{2}\, g_t \, \|\mb{v}_{t+1} - \hat{\mb{v}}_{t}\|_2^2. \label{equ:mem_loss_lookahead}
\end{align}
Intuitively, this encourages \implname\ to store information that is immediately useful for predicting the next token, similar in spirit to using short convolutions in linear-attention variants~\citep{RWKV,GLA,Mamba}.

Importantly, we apply the fast-weight update only after processing the entire chunk, so lookahead targets do not affect predictions within the same chunk. We carefully handle values at chunk boundaries. For efficiency, we drop the last predicted value in each chunk for efficiency \textit{at training time} (which results in dropping 8 tokens for a 4096-token training sequence). At inference time, a \implname~maintains a value cache that accumulates new $(\mb{v}_{t+1}, \hat{\mb{v}}_t)$ pairs. When a pre-defined update chunk size is reached (\textit{e.g.} 512 in PPL evaluation, haystack length in NIAH evaluation), \implname~consumes stored pairs to update its fast weights.

In our experiments, lookahead targets improved performance and we use it as the default unless otherwise noted.

\subsection{Inverse-Distance Weighting (IDW) Scoring} \label{app:idw}
A query--key score in PKM is typically the dot product $s_i=\mb{q}^\top K_i$.
However, a key row can increase its score by growing its magnitude, without necessarily being \emph{close} to the query in representation space.
We therefore consider inverse distance weighting (IDW,~\citealt{IDW}) as a drop-in alternative scoring function:
\begin{align}
    s^{\text{IDW}}_i = - \log \big(\epsilon + \| \mb{q} - K_i \|_2^2 \big), \label{equ:idw_score}
\end{align}
where we use $\epsilon = 10^{-3}$.
Due to the use of Euclidean distance, gradients produced by IDW scores push keys to behave as \emph{prototypes}---centroids of query clusters.
We apply IDW by replacing only the query--key scoring function inside PKM's Top-$k$ selection (for each sub-key set in product-key retrieval); all other steps of PKM retrieval remain unchanged.
Empirically, we found IDW scoring to yield better performance than dot-product scoring in our \implname\ setup.

\subsection{Target Normalization and Gradient Clipping}  \label{app:target_norm}
We z-score normalize target values $\mb{v}_t$ along the feature dimension to stabilize optimization.
Unlike standard slow-weight training, we do not clip gradients for fast-weight updates; this helps \implname\ match the scale of unbounded target values.

\subsection{Loss Reduction and Effective Step Size}
In Eq.~\ref{equ:mem_loss} we sum over tokens and features.
We avoid mean reduction so that the update magnitude does not shrink with chunk size $C$ or value dimension $d_v$.
Concretely, if one uses mean reduction over $(t,j)$, the gradient of an element $V_{i,j}$ scales by $1/(C d_v)$, which effectively changes the fast-weight step size.

\subsection{Gating-weighted Memorization}
We weight each per-token MSE by $g_t$ in Eq.~\ref{equ:mem_loss}.
Intuitively, $g_t$ measures how much the model relies on the memory output at this position, so it also serves as a natural importance weight for deciding which tokens should write more strongly. Note that we use gate $g_t$ both to interpolate outputs (Eq.~\ref{equ:fwpkm_output}) and to weight memorization updates in $\mathcal{L}_\text{mem}$ (and $\mathcal{L}_{\text{mem}}^{\text{LA}}$).
\clearpage
\section{Pseudo Code}  \label{app:pseudo_code}

\begin{algorithm}[H]
\DontPrintSemicolon
\SetKwInOut{Input}{Input}
\SetKwInOut{Output}{Output}
\SetKwInOut{Hyperparams}{Hyperparameters}

\caption{Fast-weight Product Key Memory (FwPKM) Processing}
\label{alg:fwpkm}

\Input{Hidden states $\mb{h}_{1:T} \in \mathbb{R}^{T \times d}$, \\Fast weights $\theta = \{V \in \mathbb{R}^{N \times d_v}, K^{(1)}, K^{(2)} \in \mathbb{R}^{\sqrt{N} \times d_k/2}\}$, \\Slow weights $\phi$}
\Hyperparams{Chunk size $C$, Key dim $d_k$, Value dim $d_v$, Top-K $k$}
\Output{Output sequence $\mb{o}_{1:T} \in \mathbb{R}^{T \times d_v}$}

\BlankLine
\For{$t = 1$ \KwTo $T$ \textbf{step} $C$}{
    \tcp{--- Phase 1: Forward Pass (Inference) ---}
    $\mb{h}_\text{chunk} \leftarrow \mb{h}_{t : t+C}$ \tcp*[r]{Shape: $(C, d)$}
    
    \tcp{Compute projections using Slow Weights $\phi$}
    $\mb{q} \leftarrow \mathtt{Linear}^q_\phi(\mathtt{RMSNorm}^q_\phi(\mb{h}_\text{chunk}))$ \tcp*[r]{Shape: $(C, d_k)$}
    $\mb{v} \leftarrow \mathtt{Linear}^v_\phi(\mathtt{RMSNorm}^v_\phi(\mb{h}_\text{chunk}))$ \tcp*[r]{Shape: $(C, d_v)$}
    $g \leftarrow \sigma\big(\mathtt{Linear}^g_\phi(\mathtt{RMSNorm}^g_\phi(\mb{h}_\text{chunk}))\big)$ \tcp*[r]{Shape: $(C, 1)$}

    \tcp{Split query into two halves for Product Keys}
    $\mb{q}^{(1)}, \mb{q}^{(2)} \leftarrow \mathtt{Split}(\mb{q})$ \tcp*[r]{Shapes: $(C, d_k/2)$}
    
    \tcp{Sparse Retrieval (PKM) with current Fast Weights}
    \For{$m \in \{1, 2\}$}{
        \tcp{IDW Scoring}
        $s^{(m)} \leftarrow -\log(10^{-3} + ||\mb{q}^{(m)} - K^{(m)}||_2^2)$ \tcp*[r]{Shape: $(C, \sqrt{N})$}
        \tcp{Top-k Indices}
        $\mathcal{I}^{(m)} \leftarrow \mathtt{Top-k}(s^{(m)})$ \tcp*[r]{Shape: $(C, k)$}
    }
    
    \tcp{Select final Top-k from the $k \times k$ restricted product}
    $S_\text{target} \leftarrow \{s_i^{(1)} + s_j^{(2)} \mid (i, j) \in \mathcal{I}^{(1)} \times \mathcal{I}^{(2)}\}$ \tcp*[r]{$k^2$ candidates per token}
    $\mathcal{I} \leftarrow \mathtt{Top-k}(S_\text{target})$ \tcp*[r]{Final $k$ indices per token}
    $\hat{\mb{v}} \leftarrow \sum_{(i,j) \in \mathcal{I}} \mathtt{softmax}(S_\text{target})_{(i,j)} V_{\mathtt{flat-idx}(i,j)}$ \tcp*[r]{Shape: $(C, d_v)$}
    
    \tcp{Gated Residual Output}
    $\mb{o}_\text{chunk} \leftarrow g \cdot \hat{\mb{v}} + (1-g) \cdot \mb{v}$ \tcp*[r]{Shape: $(C, d_v)$}
    
    \BlankLine \BlankLine 
    \tcp{--- Phase 2: Backward Pass for Memorization ---}
    $\mb{v}_\text{target} \leftarrow \mathtt{Shift}(\mb{v}, +1)$ \tcp*[r]{Lookahead Target: $\mb{v}_{t+1}$}
    $\mathcal{L}_\text{mem}^\text{LA} \leftarrow \sum_{\text{step}=1}^{C-1} \frac{1}{2} g_\text{step} || \mb{v}_\text{target} - \hat{\mb{v}}_\text{step} ||_2^2$\;
    
    $\nabla^\text{agg}_{V_i} \leftarrow \frac{1}{N_i^\text{read}} \nabla_{V_i} \mathcal{L}_\text{mem}$ \tcp*[r]{Aggregate gradients by slot usage}
    $V \leftarrow V - \nabla^\text{agg}_V$ \tcp*[r]{Update Values}
    
    \BlankLine \BlankLine 
    \tcp{--- Phase 3: Backward Pass for Addressing (Anti-collapsing) ---}
    \For{$m \in \{1, 2\}$}{
        $\bar{p}^{(m)} \leftarrow \mathtt{AvgSlotUsage}(\mathcal{I}^{(m)}, \text{chunk})$ \tcp*[r]{Shape: $(\sqrt{N},)$}
        $\mathcal{L}_\text{addr}^{(m)} \leftarrow -H(\bar{p}^{(m)})$ \tcp*[r]{Maximize Entropy}
        $K^{(m)} \leftarrow K^{(m)} - \nabla_{K^{(m)}} \mathcal{L}_\text{addr}^{(m)}$ \tcp*[r]{Update Keys}
    }
}
\KwRet{$\mb{o}_{1:T}$}
\end{algorithm}
\clearpage
\section{Detailed Experiment Settings} \label{app:exp_setting}

\begin{table}[h]
    \centering
    \caption{\footnotesize{Modeling and training hyper-parameters used in the experiments.} \label{tab:modeling-hyperparams}}
    
    % --- First Subtable (Left Column) ---
    \begin{subtable}[t]{0.25\textwidth}
        \centering
        \begin{tabular}{lr}
            \toprule
            \textbf{Hyperparameter} & \textbf{Value} \\
            \midrule
            \multicolumn{2}{l}{\textit{General}} \\
            \quad vocab. size & 32000 \\
            \quad \# layers & 12 \\
            \quad hidden dim & 768 \\
            \quad RMS norm $\epsilon$ & $0.00001$ \\
            \midrule
            \multicolumn{2}{l}{\textit{Attention}} \\
            \quad head dim & 64 \\
            \quad \# query heads & 12 \\
            \quad \# k/v heads & 4 \\
            \midrule
            \multicolumn{2}{l}{\textit{GDN}} \\
            \quad conv. size & 4 \\
            \quad head dim & 64 \\
            \quad \# heads & 8 \\
            \midrule
            \multicolumn{2}{l}{\textit{PKM}} \\
            \quad key dim & 512 \\
            \quad value dim & 512 \\
            \quad \# heads & 4 \\
            \quad Top-$K$ & 32 \\
            \quad \# slots & $512^2$ \\
            \bottomrule
        \end{tabular}
    \end{subtable}
    \hfill % Adds space between the two tables
    % --- Second Subtable (Right Column) ---
    \begin{subtable}[t]{0.25\textwidth}
        \centering
        \begin{tabular}{lr}
            \toprule
            \textbf{Hyperparameter} & \textbf{Value} \\
            \midrule
            \multicolumn{2}{l}{\textit{FwPKM}} \\
            \quad key dim & 512 \\
            \quad value dim & 512 \\
            \quad \# heads & 1 \\
            \quad Top-$K$ & 8 \\
            \quad \# slots & $512^2$ \\
            \quad chunk size & 512 \\
            \midrule
            \multicolumn{2}{l}{\textit{FwMLP}} \\
            \quad input dim & 512 \\
            \quad hidden dim & 2304 \\
            \quad output dim & 512 \\
            \quad LR $\eta$ & 0.1 \\
            \midrule
            \multicolumn{2}{l}{\textit{LaCT}} \\
            \quad sliding window size & 2048 \\
            \quad TTT chunk size & 512 \\
            \bottomrule
        \end{tabular}
    \end{subtable}
    \hfill % Adds space between the two tables
    % --- Second Subtable (Right Column) ---
    \begin{subtable}[t]{0.35\textwidth}
        \centering
        \begin{tabular}{lr}
            \toprule
            \textbf{Hyperparameter} & \textbf{Value} \\
            \midrule
            \multicolumn{2}{l}{\textit{Training}} \\
            \quad max. LR & $0.001$ \\
            \quad min. LR & $0.0001$ \\
            \quad global batch size & 128 \\
            \quad micro batch size & 8 \\
            \quad \# warmup steps & 100 \\
            \quad \# total steps & 20000 \\
            \quad weight decay & 0.1 \\
            \bottomrule
        \end{tabular}
    \end{subtable}
\end{table}

\paragraph{\textbf{Training}} Every training experiment is conducted on 4 H100 GPUs. The \implname~module runs under \texttt{Float32} precision, other components are converted to \texttt{BFloat16} unless specified by the FlashAttention~\citep{FlashAttention,FlashAttention2} and the FLA~\citep{FLA} kernels.

\paragraph{\textbf{Evaluation}} Every evaluation experiment is conducted on 1 H100 GPU. We apply the YaRN method~\citep{YaRN} to adapt RoPE embeddings to the maximum sequence length in the experiments where sample length exceeds 4096.

\paragraph{\textbf{Baseline model - \texttt{FwMLP}}} To rule out the impact of implementation details unrelated to the PKM architecture, we propose a baseline model that replaces the PKM in \implname~with a SwiGLU-MLP that maintains three fast-weight matrices (and their biases) for up, gating, and down projection. The baseline, denoted as \texttt{FwMLP}, updates its fast weights by minimizing the MSE loss between its predicted values and target lookahead values at a chunk level. Due to its dense nature, we reduce the MSE losses in a chunk by averaging over both sample and feature dimensions, and we do not apply the loss aggregation and gradient shaping techniques mentioned in Section~\ref{ssec:mem_optim_main}. For the same reason, addressing optimization in Section~\ref{ssec:addr_optim} is irrelevant too.

\paragraph{\textbf{Baseline model - \texttt{LaCT}}} We adopt the official implementation of LaCT~\citep{TTTPlus}\footnote{\url{https://github.com/a1600012888/LaCT/tree/main/lact_llm}} as a strong TTT~\citep{TTT} baseline. The LaCT architecture consists of a sliding window attention, a fast-weight SwiGLU MLP, and a slow-weight SwiGLU MLP in every layer. The fast weights are optimized to minimize a dot product loss via SGD with momentum or Muon~\citep{muon}, and we found SGD with momentum achieves lower training PPL in our experiments. LaCT uses data-dependent learning rate and L2 weight normalization to improve memorization and retention. We compared several configurations of sliding window size (W) and update chunk size (C). Among $512C+512W$, $512C+2048W$, $512C+4096W$, and $2048C+2048W$, the best model is $512C+2048W$.

One notable difference between LaCT (or more generally TTT) and FwMLP/FwPKM is that LaCT/TTT maintains an individual set of fast weights for each sequence in a mini batch, while FwMLP/FwPKM uses a shared set of fast weights for all sequences.

\clearpage

\section{Ablation Study} \label{app:ablation}

To understand the influence of techniques proposed in Section~\ref{sec:fwpkm}, we conduct ablation experiments based on the \texttt{GDN+SWA | PKM6+FwPKM@2,10} model. The following variants are trained and evaluated using the same pipelines.
\begin{itemize}[nosep]
    \item ``\texttt{w/ $1$ head $\times$ Top-$32$}'' uses a different Top-$K$ setting as the name suggests.
    \item ``\texttt{w/ $4$ heads $\times$ Top-$8$}'' uses a different multi-head setting as the name suggests.
    \item ``\texttt{w/o value norm}'' does NOT z-score normalize target values.
    \item ``\texttt{w/o addr loss}'' does NOT use the marginal entropy loss to update key matrices, instead it uses the MSE loss to update both the keys and value matrices.
    \item ``\texttt{w/o gating}'' does NOT use $g_t$.
    \item ``\texttt{w/o loss weight}'' uses $g_t$ but does NOT weight MSE loss with it.
    \item ``\texttt{w/o lookahead}'' does NOT use lookahead values as MSE targets.
\end{itemize}
In addition, we found that replacing IDW score with dot-product score often results in memory collapsing and loss divergence so do not include it in the ablation study.

\subsection{Evaluation}
\begin{figure}[h!]
    \begin{center}
        \includegraphics[width=0.98\textwidth]{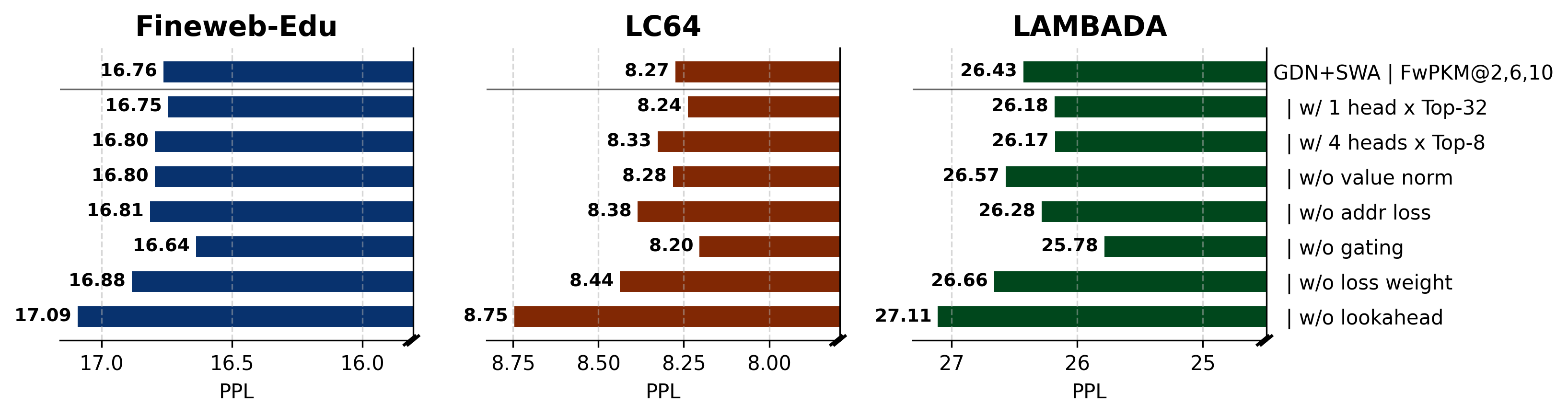}
    \end{center}
    \caption{\footnotesize{Ablation study: Perplexity on Fineweb-Edu, LC64, and LAMBADA. We use bar colors to help distinguish between models with different types of memory components.} \label{fig:eval_ppl_ablation}}
\end{figure}

\begin{figure}[]
    \begin{center}
        \includegraphics[width=0.95\textwidth]{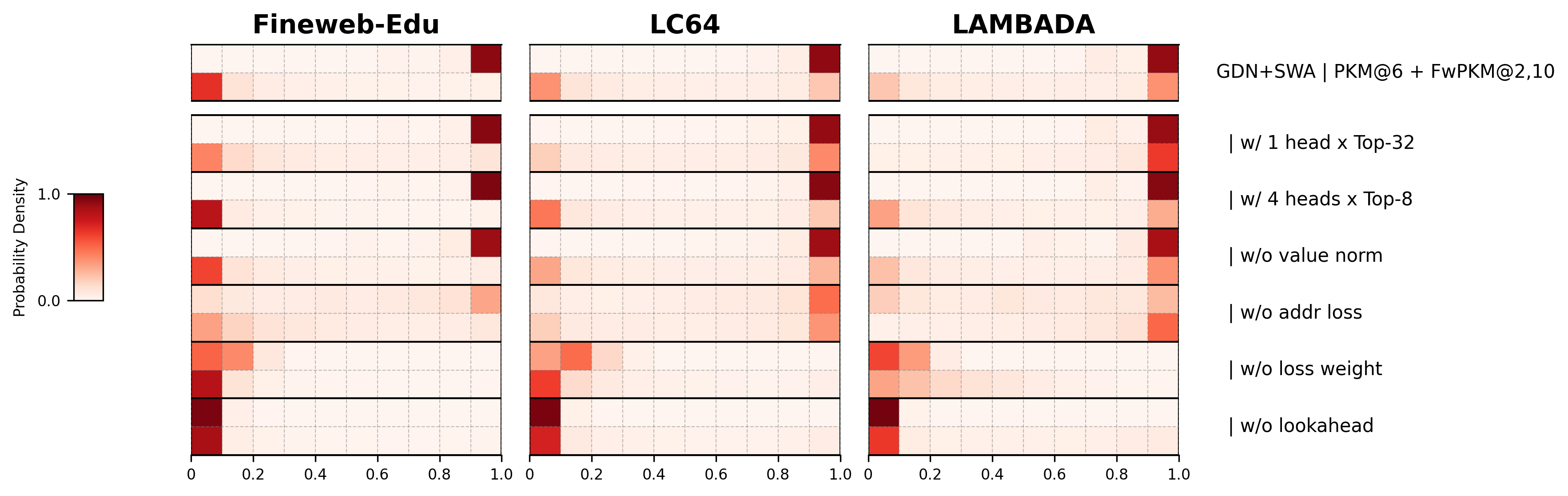}
    \end{center}
    \caption{\footnotesize{Ablation study: \implname~gating value distribution on Fineweb-Edu, LC64, and LAMBADA test sets. Each row represents one \implname~layer. The probability densities are categorized into 10 bins. A darker color suggests a higher probability density of a bin.}}
    \label{fig:eval_gate_ablation}
\end{figure}

\begin{figure}[]
    \begin{center}
        \includegraphics[width=1.0\textwidth]{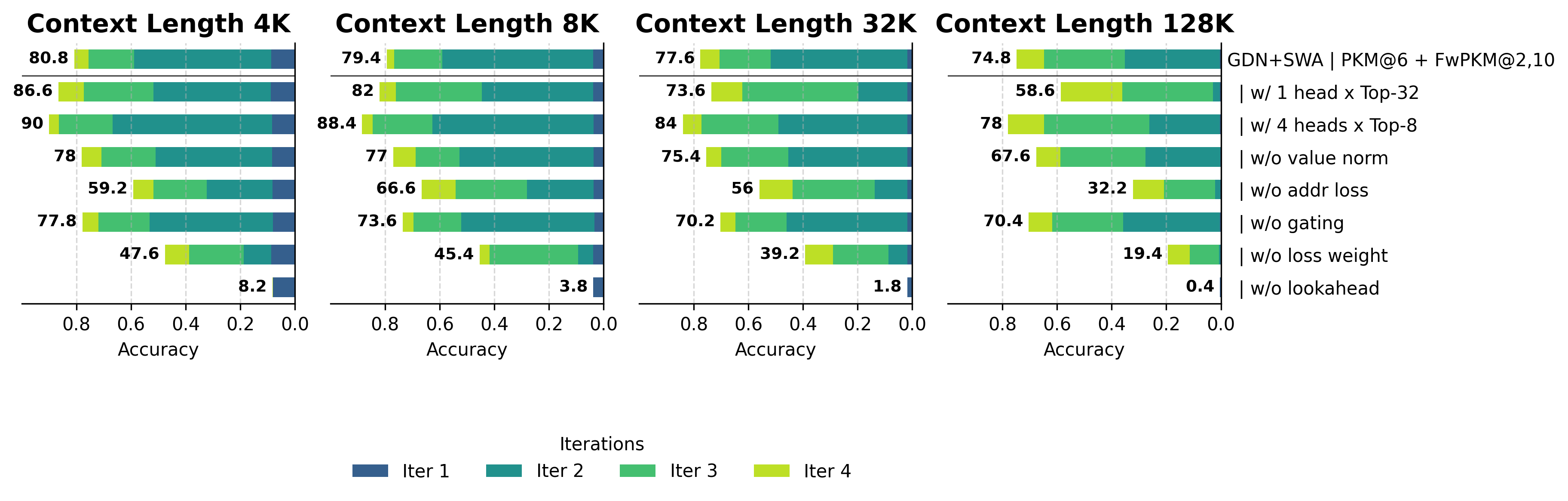}
    \end{center}
    \caption{\footnotesize{Ablation study: Stacked bar plots for NIAH accuracy results on 4K-/8K-/32K-/128K-length test sets. Each stacked bar shows the accuracies of $\{1,2,3,4\}$-iter NIAH evaluations.} \label{fig:eval_niah_ablation}}
\end{figure}

\begin{figure}[]
    \begin{center}
        \includegraphics[width=1.0\textwidth]{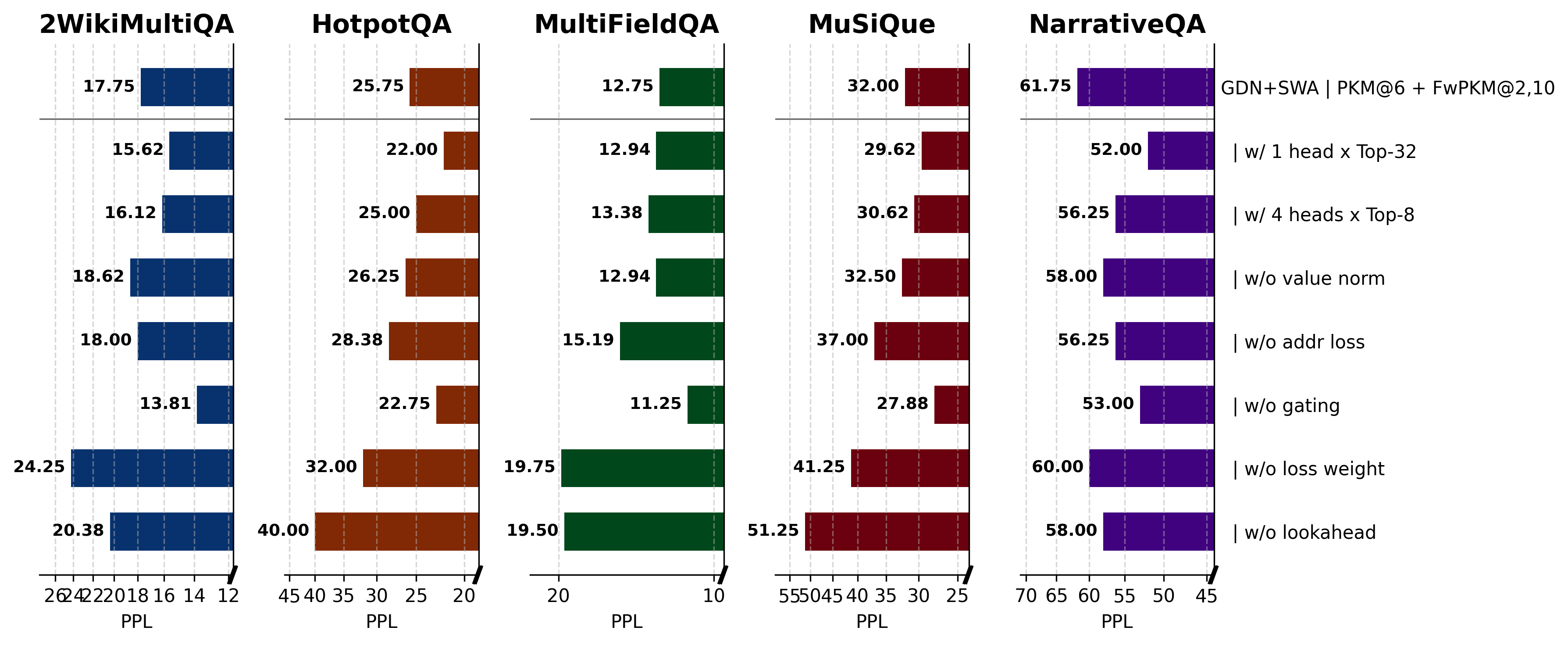}
    \end{center}
    \caption{\footnotesize{Ablation study: Perplexity on five Longbench tasks. We use bar colors to help distinguish between models with different types of memory components.} \label{fig:eval_longbench_ablation}}
\end{figure}

\begin{figure}[]
    \begin{center}
        \includegraphics[width=0.8\textwidth]{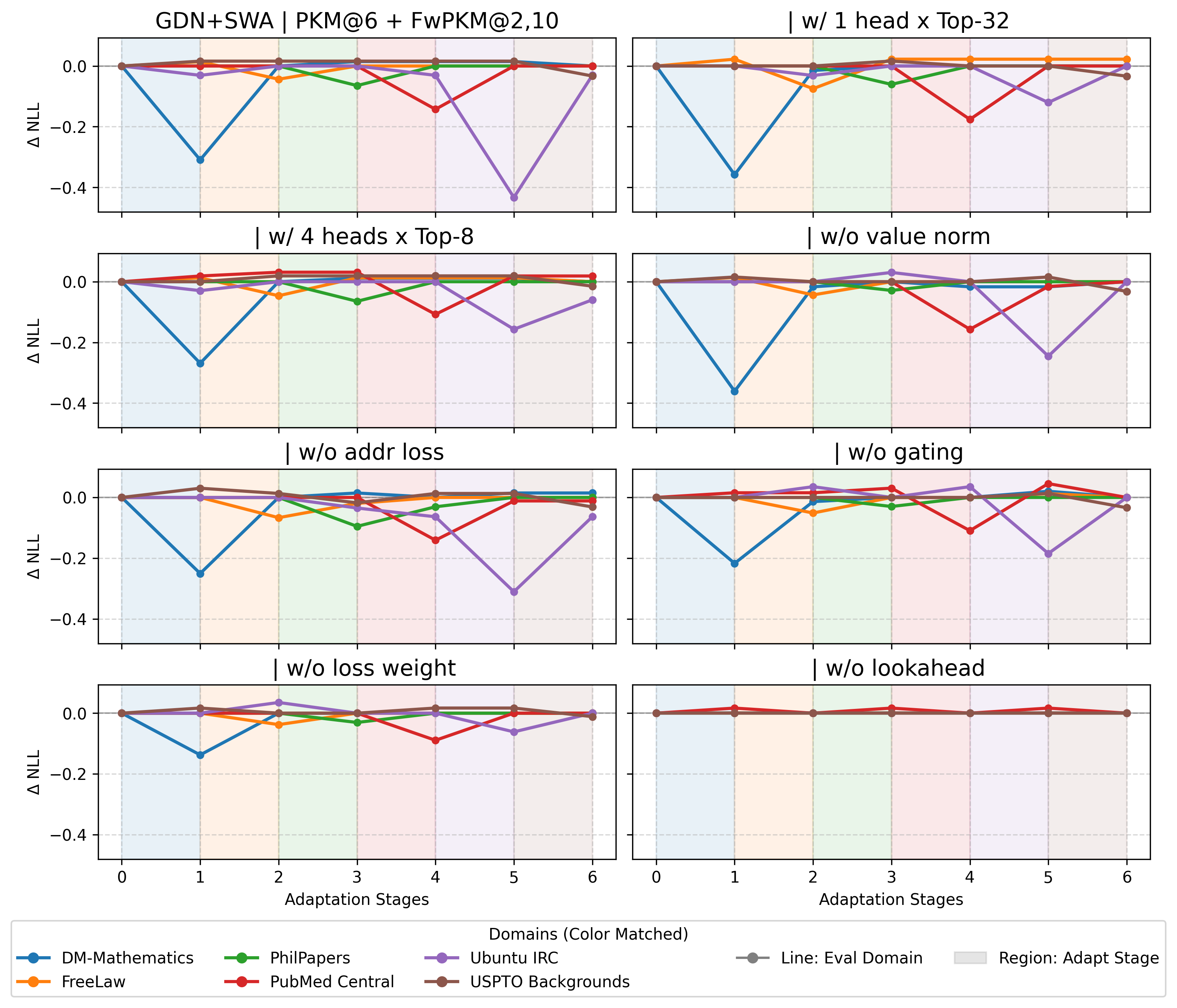}
    \end{center}
    \caption{\footnotesize{Ablation study: Change of negative log-likelihood ($\Delta$NLL) on six Pile domain datasets before and after 6 adaptation stages of test-time learning on each domain dataset.} \label{fig:eval_pile_domain_ablation}}
\end{figure}

As shown in Figures~\ref{fig:eval_ppl_ablation}, \ref{fig:eval_gate_ablation}, \ref{fig:eval_niah_ablation}, \ref{fig:eval_longbench_ablation}, and \ref{fig:eval_pile_domain_ablation}, we conduct the same evaluation experiments from Section~\ref{sec:experiments} to ablated models. a) Notably, removing lookahead values yields the most significant harm to model performance. Many techniques bring slight PPL improvement, but lead to less healthy memory utility and subsequently worse NIAH accuracies to different extents. b) Addressing loss and weighting MSE loss with gating values are important for NIAH retrieval accuracy. c) A larger effective top-$k$ slightly improves performance in some tasks, but the default setting of $1$ head $\times$ $8$ slots strikes a sweet spot between task performance and compute efficiency (See computation cost of ablated models in Table~\ref{tab:model_comparison_ablation}).

% \clearpage

\begin{table}[]
\caption{Ablation study: Comparison of model size and computation cost.}
\label{tab:model_comparison_ablation}
\begin{center}
\begin{small}
\begin{tabular}{lrrrrr}
\toprule
\multirow{2}{*}{\textsc{Model}} & \multicolumn{2}{c}{Parameters (M)} & \multirow{2}{*}{FLOPs (T)} & \multirow{2}{*}{FLOPS (T/sec.)} & \multirow{2}{*}{Samples/sec.} \\
\cmidrule(lr){2-3}
 & Total & Active & & & \\
 \midrule
\texttt{GDN+SWA | PKM@6 + FwPKM@2,10} & 516.31 & 113.73 & 22.24 & 58.55 & 21.06 \\
\texttt{  | w/ 1 head x Top-32} & 516.31 & 113.76 & 22.24 & 58.28 & 20.96 \\
\texttt{  | w/ 4 heads x Top-8} & 520.24 & 117.69 & 22.71 & 48.47 & 17.08 \\
\texttt{  | w/o value norm} & 516.31 & 113.73 & 22.24 & 65.03 & 23.39 \\
\texttt{  | w/o addr loss} & 516.31 & 113.73 & 22.24 & 65.91 & 23.70 \\
\texttt{  | w/o gating} & 516.30 & 113.72 & 22.24 & 61.30 & 22.05 \\
\texttt{  | w/o loss weight} & 516.31 & 113.73 & 22.24 & 61.06 & 21.96 \\
\texttt{  | w/o lookahead} & 516.31 & 113.73 & 22.24 & 58.70 & 21.11 \\
\bottomrule
\end{tabular}
\end{small}
\end{center}
\end{table}

\clearpage
\subsection{Freezing Fast Weights}  \label{app:freeze_fw}

\begin{figure}[h]
    \begin{center}
        \includegraphics[width=0.98\textwidth]{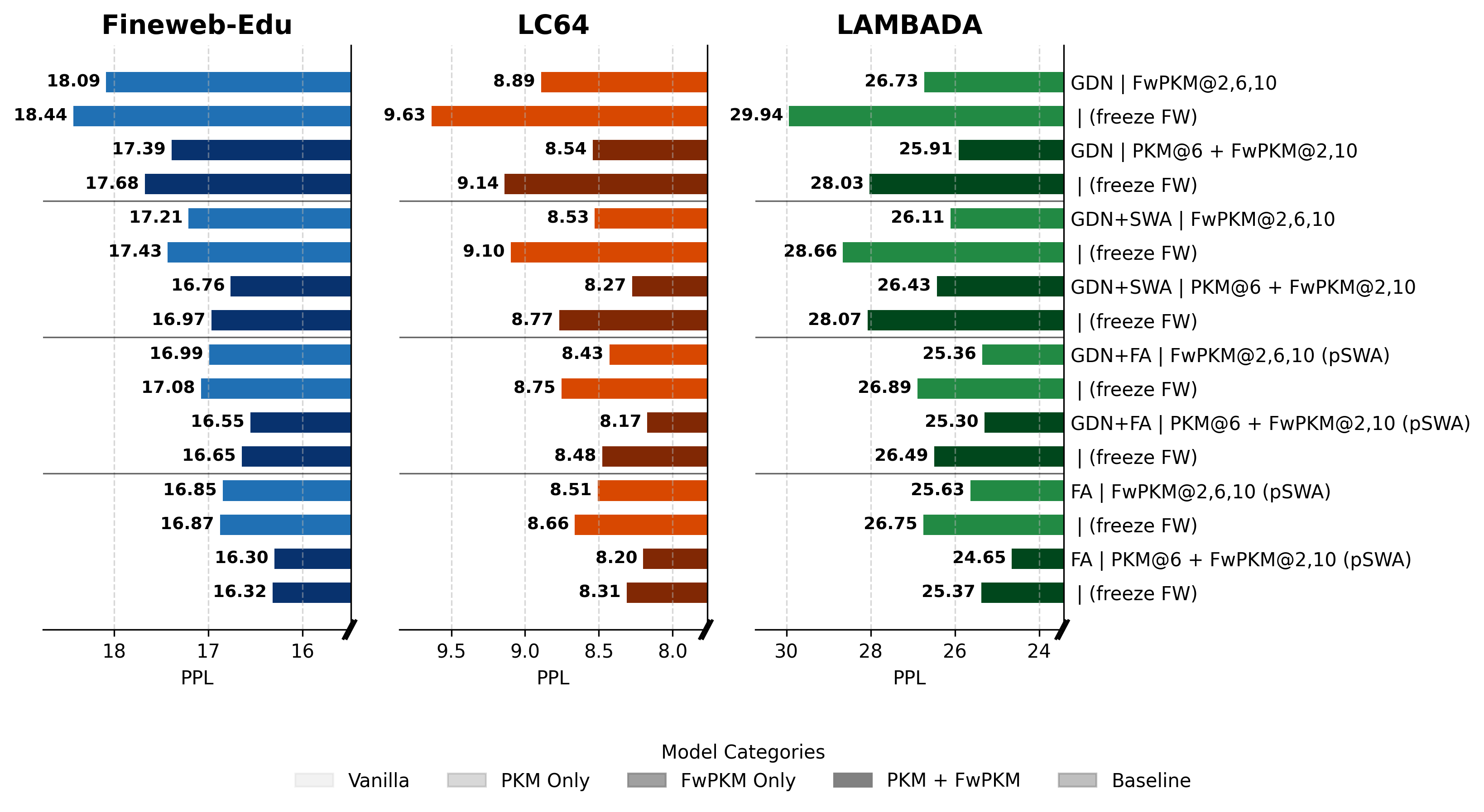}
    \end{center}
    \caption{\footnotesize{Ablation study: Perplexity on Fineweb-Edu, LC64, and LAMBADA. Unlike previous experiments, the fast-weight parameters in \implname~are frozen. We use bar colors to help distinguish between models with different types of memory components.} \label{fig:eval_ppl_freeze_fw}}
\end{figure}

To consolidate the evidence that \implname~contributes to long-context language modeling, we conduct extra PPL evaluation experiments as in Section~\ref{ssec:ppl_eval} but \textit{freezing} \implname's fast-weight parameters (\textit{i.e.} value matrix $V$ and key matrices $K^{(1)}, K^{(2)}$). Figure~\ref{fig:eval_ppl_freeze_fw} shows that freezing fast weights significantly increase perplexity on long-context datasets LC64 and LAMBADA while Fineweb-Edu is less affected, further supporting conclusions in Section~\ref{ssec:ppl_eval}. 

\clearpage
\subsection{Addressing Metrics}
The design choices have different impact on the evenness of slot usage, and we use three addressing metrics to characterize it. Recall that we define $N^\text{read}_i$ to be the ``row contribution'', \textit{i.e.} the number of times a value row $V_i$ is accessed.
\begin{itemize}[nosep]
    \item \textbf{Collision rate} is the ratio of \textit{competitive} slot accesses: $\sum_{N^\text{read}_i > 1} N^\text{read}_i / N^\text{total}$, where $N^\text{total}$ is the total number of slot accesses.
    \item \textbf{Coverage rate} is the ratio of accessed slots: $\sum_{N^\text{read}_i > 0} 1$ / N.
    \item \textbf{KLD} is the Kullback–Leibler divergence between the average slot use distribution and an uniform distribution:
    \begin{align*}
        p^\text{read}_i &= N^\text{read}_i / \sum_i^N N^\text{read}_i \\
        \text{KLD} &= \sum_i^N p^\text{read}_i \log \frac{p^\text{read}_i}{1/N}.
    \end{align*}
\end{itemize}

For each evaluation dataset and each metric, we calculate the metric every 32K tokens, which gives us 8M / 32K = 256 numbers and we take their average. We choose 32K tokens because a \texttt{GDN+SWA | PKM@6+FwPKM@2,10} model selects 8 slots per token and perfectly balanced slot selection would cover 32K$\times$8=256K slots, which is the exact number of \implname~slots. We report the metrics averaged across two \implname~layers.

As shown in Table~\ref{tab:addr_metrics}, memory slot use is more balanced on long-context datasets LAMBADA and LC64 but still far from uniform distribution. Increasing head number or Top-$K$ can improve the evenness but also increases collision rate. Ablating the addressing loss leads to a concentrated slot access pattern. 

Unhealthy slot access statistics such as in the ``\texttt{w/o addr loss}'' model are paired with degraded performance in certain tasks (\textit{e.g.} NIAH). If we compare models of effectively Top-32 slot activations (\textit{i.e.} ``\texttt{w/ 1 head x Top-32}'' and ``\texttt{w/ 4 heads x Top-8}'') against models of effectively Top-8 slot activations, despite the former group having higher coverage and lower KLD, their differences in evaluation experiments (Figures~\ref{fig:eval_ppl_ablation}, \ref{fig:eval_gate_ablation}, \ref{fig:eval_niah_ablation}, \ref{fig:eval_longbench_ablation}, and \ref{fig:eval_pile_domain_ablation}) do not correlate with the difference in slot access statistics. The results might be affected by compound factors such as the Top-$k$ mechanism per se, but we believe it is important to mitigate severely unbalanced memory slot access, for example by minimizing the addressing loss.

\begin{table*}[h!]
\caption{\footnotesize{Memory addressing metrics. Coll. -- collision rate (\%). Cov. -- coverage rate (\%). KLD -- KLD against uniform distribution. Darker colors indicate better performance (\textit{i.e.} low collision, high coverage, low KLD).} \label{tab:addr_metrics}}
\begin{center}
\begin{adjustbox}{width=\textwidth,center}
\begin{tabular}{lccc|ccc|ccc}
\toprule
\multirow{2}{*}{Model} & \multicolumn{3}{c|}{\textbf{Fineweb-Edu}} & \multicolumn{3}{c|}{\textbf{LAMBADA}} & \multicolumn{3}{c}{\textbf{LC64}} \\
 & Coll. $\downarrow$ & Cov. $\uparrow$ & KLD $\downarrow$ & Coll .$\downarrow$ & Cov .$\uparrow$ & KLD $\downarrow$ & Coll .$\downarrow$ & Cov .$\uparrow$ & KLD $\downarrow$ \\
\midrule
\texttt{GDN+SWA | PKM@6 + FwPKM@2,10} & \cellcolor[HTML]{FB7C5C}\textcolor{black}{55.7} & \cellcolor[HTML]{DDF2D8}\textcolor{black}{15.8} & \cellcolor[HTML]{4493C7}\textcolor{black}{2.2} & \cellcolor[HTML]{FDD3C1}\textcolor{black}{83.1} & \cellcolor[HTML]{B1E0AB}\textcolor{black}{32.1} & \cellcolor[HTML]{1A68AE}\textcolor{white}{1.9} & \cellcolor[HTML]{FEDACA}\textcolor{black}{85.2} & \cellcolor[HTML]{BDE5B6}\textcolor{black}{28.4} & \cellcolor[HTML]{084E98}\textcolor{white}{2.1} \\
\texttt{| w/ 1 head x Top-32} & \cellcolor[HTML]{FDC5AE}\textcolor{black}{78.1} & \cellcolor[HTML]{92D28F}\textcolor{black}{41.7} & \cellcolor[HTML]{08306B}\textcolor{white}{1.3} & \cellcolor[HTML]{FFEDE5}\textcolor{black}{95.3} & \cellcolor[HTML]{3DA65A}\textcolor{white}{64.1} & \cellcolor[HTML]{08306B}\textcolor{white}{1.3} & \cellcolor[HTML]{FFEEE6}\textcolor{black}{95.4} & \cellcolor[HTML]{60BA6C}\textcolor{black}{55.0} & \cellcolor[HTML]{084285}\textcolor{white}{1.9} \\
\texttt{| w/ 4 heads x Top-8} & \cellcolor[HTML]{FDC5AE}\textcolor{black}{78.2} & \cellcolor[HTML]{94D390}\textcolor{black}{41.4} & \cellcolor[HTML]{083471}\textcolor{white}{1.4} & \cellcolor[HTML]{FFEEE6}\textcolor{black}{95.5} & \cellcolor[HTML]{48AE60}\textcolor{black}{60.8} & \cellcolor[HTML]{083D7F}\textcolor{white}{1.4} & \cellcolor[HTML]{FFEEE6}\textcolor{black}{95.3} & \cellcolor[HTML]{48AE60}\textcolor{black}{60.9} & \cellcolor[HTML]{08306B}\textcolor{white}{1.7} \\
\texttt{| w/o value norm} & \cellcolor[HTML]{FB7C5C}\textcolor{black}{55.6} & \cellcolor[HTML]{DDF2D8}\textcolor{black}{15.8} & \cellcolor[HTML]{4191C6}\textcolor{white}{2.2} & \cellcolor[HTML]{FDD3C1}\textcolor{black}{83.1} & \cellcolor[HTML]{B1E0AB}\textcolor{black}{32.1} & \cellcolor[HTML]{1967AD}\textcolor{white}{1.8} & \cellcolor[HTML]{FED9C9}\textcolor{black}{84.9} & \cellcolor[HTML]{BCE4B5}\textcolor{black}{28.9} & \cellcolor[HTML]{084B93}\textcolor{white}{2.1} \\
\texttt{| w/o addr loss} & \cellcolor[HTML]{FED8C7}\textcolor{black}{84.6} & \cellcolor[HTML]{EDF8E9}\textcolor{black}{7.2} & \cellcolor[HTML]{F7FBFF}\textcolor{black}{3.6} & \cellcolor[HTML]{FFEDE5}\textcolor{black}{94.9} & \cellcolor[HTML]{E6F5E1}\textcolor{black}{12.0} & \cellcolor[HTML]{F7FBFF}\textcolor{black}{3.9} & \cellcolor[HTML]{FFEFE8}\textcolor{black}{96.4} & \cellcolor[HTML]{ECF8E8}\textcolor{black}{8.0} & \cellcolor[HTML]{F7FBFF}\textcolor{black}{5.4} \\
\texttt{| w/o gating} & \cellcolor[HTML]{FB7B5B}\textcolor{black}{55.2} & \cellcolor[HTML]{DDF2D8}\textcolor{black}{16.0} & \cellcolor[HTML]{3F8FC5}\textcolor{white}{2.2} & \cellcolor[HTML]{FDD3C1}\textcolor{black}{83.1} & \cellcolor[HTML]{B1E0AB}\textcolor{black}{32.1} & \cellcolor[HTML]{1A68AE}\textcolor{white}{1.9} & \cellcolor[HTML]{FDD7C6}\textcolor{black}{84.3} & \cellcolor[HTML]{B8E3B2}\textcolor{black}{29.9} & \cellcolor[HTML]{08488E}\textcolor{white}{2.0} \\
\texttt{| w/o loss weight} & \cellcolor[HTML]{FC8161}\textcolor{black}{57.3} & \cellcolor[HTML]{DEF2D9}\textcolor{black}{15.5} & \cellcolor[HTML]{4594C7}\textcolor{black}{2.2} & \cellcolor[HTML]{FDD5C4}\textcolor{black}{83.8} & \cellcolor[HTML]{B5E1AE}\textcolor{black}{31.2} & \cellcolor[HTML]{1C6AB0}\textcolor{white}{1.9} & \cellcolor[HTML]{FEDBCC}\textcolor{black}{85.6} & \cellcolor[HTML]{BEE5B8}\textcolor{black}{28.1} & \cellcolor[HTML]{084C95}\textcolor{white}{2.1} \\
\texttt{| w/o lookahead} & \cellcolor[HTML]{FB7252}\textcolor{black}{52.5} & \cellcolor[HTML]{DBF1D6}\textcolor{black}{16.6} & \cellcolor[HTML]{3787C0}\textcolor{white}{2.1} & \cellcolor[HTML]{FDD1BE}\textcolor{black}{82.4} & \cellcolor[HTML]{AFDFA8}\textcolor{black}{33.1} & \cellcolor[HTML]{1663AA}\textcolor{white}{1.8} & \cellcolor[HTML]{FDD2BF}\textcolor{black}{82.5} & \cellcolor[HTML]{B0DFAA}\textcolor{black}{32.8} & \cellcolor[HTML]{083B7C}\textcolor{white}{1.8} \\\bottomrule
\end{tabular}
\end{adjustbox}
\end{center}
\end{table*}

\clearpage
\section{More Visualization Examples}

\subsection{Improving Perplexity over Sequence Boundary} \label{app:ppl_diff_over_seq_boundary}
In long-context datasets LC64 and LAMBADA, a document can span over multiple 4K-token sequences. The ability of carrying over memory states across sequence boundaries makes \implname~models excel at reducing the PPL of tokens that require cross-sequence dependency.

\begin{figure*}[h!]
    \begin{center}
        \includegraphics[width=1.0\textwidth]{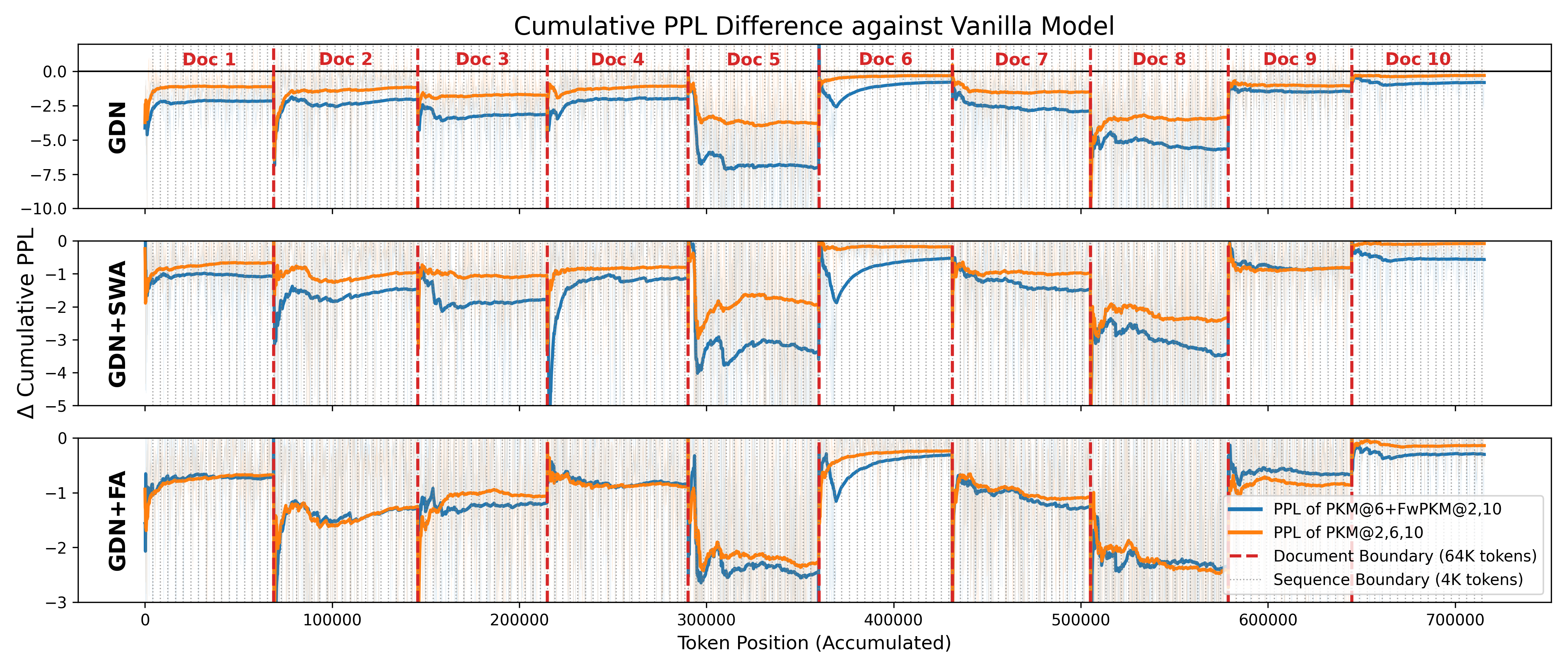}
    \end{center}
    \caption{\footnotesize{PPL reduction of ``\texttt{*|PKM@6+FwPKM@2,10}'' and ``\texttt{*|PKM@2,6,10}'' from vanilla \texttt{GDN}, \texttt{GDN+SWA}, and \texttt{GDN+FA} models at different token positions.} \label{fig:ppl_diff_over_seq_boundary}}
\end{figure*}

In Figure~\ref{fig:ppl_diff_over_seq_boundary}, we demonstrate the advantage of \implname~models (\texttt{*|PKM@6+FwPKM@2,10}) by showing their PPL reduction from vanilla counterparts (\texttt{GDN}, \texttt{GDN+SWA}, \texttt{GDN+FA}) on the first ten LC64 documents used in Section \ref{ssec:ppl_eval}. Within each document, we calculate the average of PPLs cumulated from the beginning of the document. Then we plot the reduction of cumulative PPL every 128 tokens.
The orange lines denote the cumulative PPL difference between ``\texttt{*|PKM@6+FwPKM@2,10}'' and their vanilla counterparts, and the blue lines show the PPL difference between ``\texttt{*|PKM@2,6,10}'' and vanilla counterparts as reference.

While the PKM-only variants show significant PPL reduction over the vanilla models, the PPL reduction curves are relatively smooth, suggesting constant advantage throughout each document. On the other hand, the \implname~models have more jagged curves. We hypothesize that these ``PPL drops'' are caused by \implname~exploiting information from previous sequences, which is impossible for the vanilla and PKM-only models.

\subsection{Extra Selective Gating Example}
\begin{figure*}
    \begin{center}
        \includegraphics[width=0.85\textwidth]{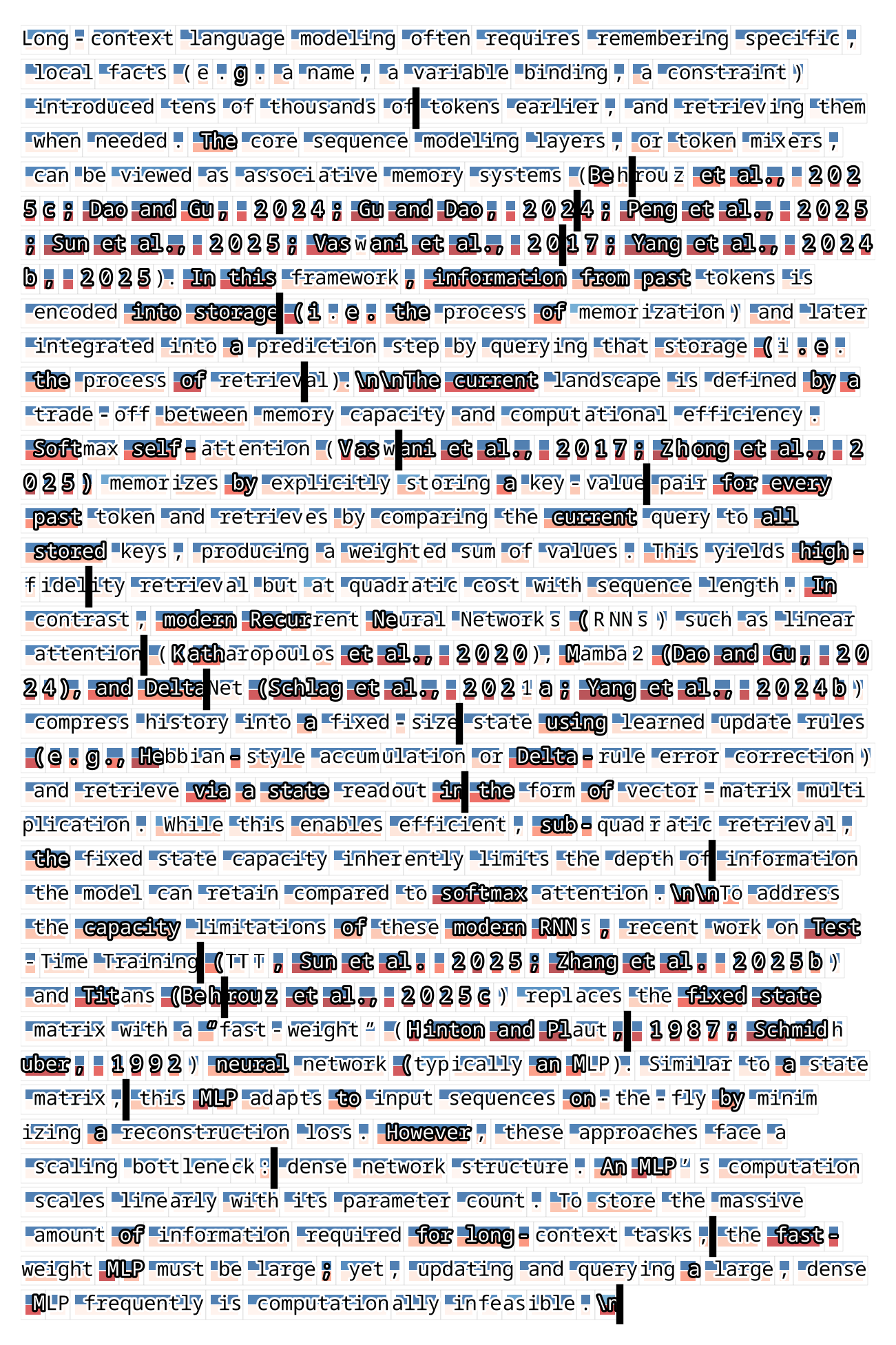}
    \end{center}
    \caption{\footnotesize{\texttt{GDN+SWA|PKM@6+FwPKM@2,10}'s \implname~gating values on tokens from the Introduction section. The blue color in the background denotes the gating intensity of the \implname~at layer 2 and the red color denotes the \implname~at layer 10. A darker color represents higher intensity. Black vertical lines show positions where \implname's fast weights are updated, which is every 32 tokens as we specified for this visualization example.}}
    \label{fig:gating_example_intro}
\end{figure*}

Figure~\ref{fig:gating_example_intro} shows the token-level \implname~gating values for the Introduction section of this paper.

\end{document}